\documentclass[10pt,twocolumn,letterpaper]{article}

\usepackage[pagenumbers]{cvpr} 

\usepackage{graphicx}
\usepackage{amsmath}
\usepackage{amssymb}
\usepackage{booktabs}

\usepackage{times}
\usepackage{epsfig}
\usepackage{graphicx}
\usepackage{amsmath}
\usepackage{amssymb}
\usepackage{color}
\usepackage{subcaption}
\usepackage{mathtools}
\usepackage{paralist} 
\usepackage{enumitem}
\usepackage{colortbl}
\usepackage{nicefrac}
\usepackage{multirow}
\usepackage[table,dvipsnames]{xcolor}
\usepackage{booktabs}
\usepackage{makecell}
\usepackage{gensymb}
\usepackage{microtype}
\usepackage{cancel}
\usepackage{wasysym}

\usepackage[numbers]{natbib}

\definecolor{yellow}{rgb}{1,1, 0.7}
\definecolor{orange}{rgb}{1, 0.8, 0.6}
\definecolor{red}{rgb}{1, 0.6, 0.6}
\definecolor{tangerine}{rgb}{0.95, 0.52, 0.}

\definecolor{darkyellow}{rgb}{0.8, 0.8, 0.5}
\definecolor{darkred}{rgb}{0.7, 0.3, 0.3}
\definecolor{darkgreen}{rgb}{0.3, 0.7, 0.3}
\definecolor{blue}{rgb}{0, 0, 1.0}
\definecolor{green}{rgb}{0, 1.0, 0}
\definecolor{pink}{rgb}{1, 0.4, 0.7}
\definecolor{lightblue}{rgb}{0.2, 0.5, 0.9}
\definecolor{todored}{rgb}{1, 0, 0}

\let\originalleft\left
\let\originalright\right
\renewcommand{\left}{\mathopen{}\mathclose\bgroup\originalleft}
\renewcommand{\right}{\aftergroup\egroup\originalright}

\DeclareMathSymbol{\shortminus}{\mathbin}{AMSa}{"39}

\def\eg{e.g\onedot} 
\def\ie{i.e\onedot} 
 
\def\etc{etc\onedot} \def\vs{vs\onedot}
\def\wrt{w.r.t\onedot}

\newcommand{\model}{DiffusionRig\xspace}

\newcommand{\myparagraph}[1]{\vspace{-4pt}\paragraph{#1}}

\newcommand\blfootnote[1]{%
  \begingroup
  \renewcommand\thefootnote{}\footnote{#1}%
  \addtocounter{footnote}{-1}%
  \endgroup
}

\usepackage[pagebackref,breaklinks,colorlinks]{hyperref}

\usepackage[capitalize]{cleveref}
\crefname{section}{Sec.}{Secs.}
\Crefname{section}{Section}{Sections}
\Crefname{table}{Table}{Tables}
\crefname{table}{Tab.}{Tabs.}

\def\cvprPaperID{3995} 
\def\confName{CVPR}
\def\confYear{2023}

\begin{document}

\title{DiffusionRig: Learning Personalized Priors for Facial Appearance Editing}

\author{Zheng Ding$^{1\dagger}$, Xuaner Zhang$^{2}$, Zhihao Xia$^{2}$, Lars Jebe$^{2}$, Zhuowen Tu$^{1}$, Xiuming Zhang$^{2}$\\
$^{1}$UC San Diego \quad \quad $^{2}$Adobe
}

\twocolumn[{%
\renewcommand\twocolumn[1][]{#1}%
\maketitle

\vspace{-3em}
\begin{center}
    \centering
    \captionsetup{type=figure}
    \includegraphics[width=\textwidth]{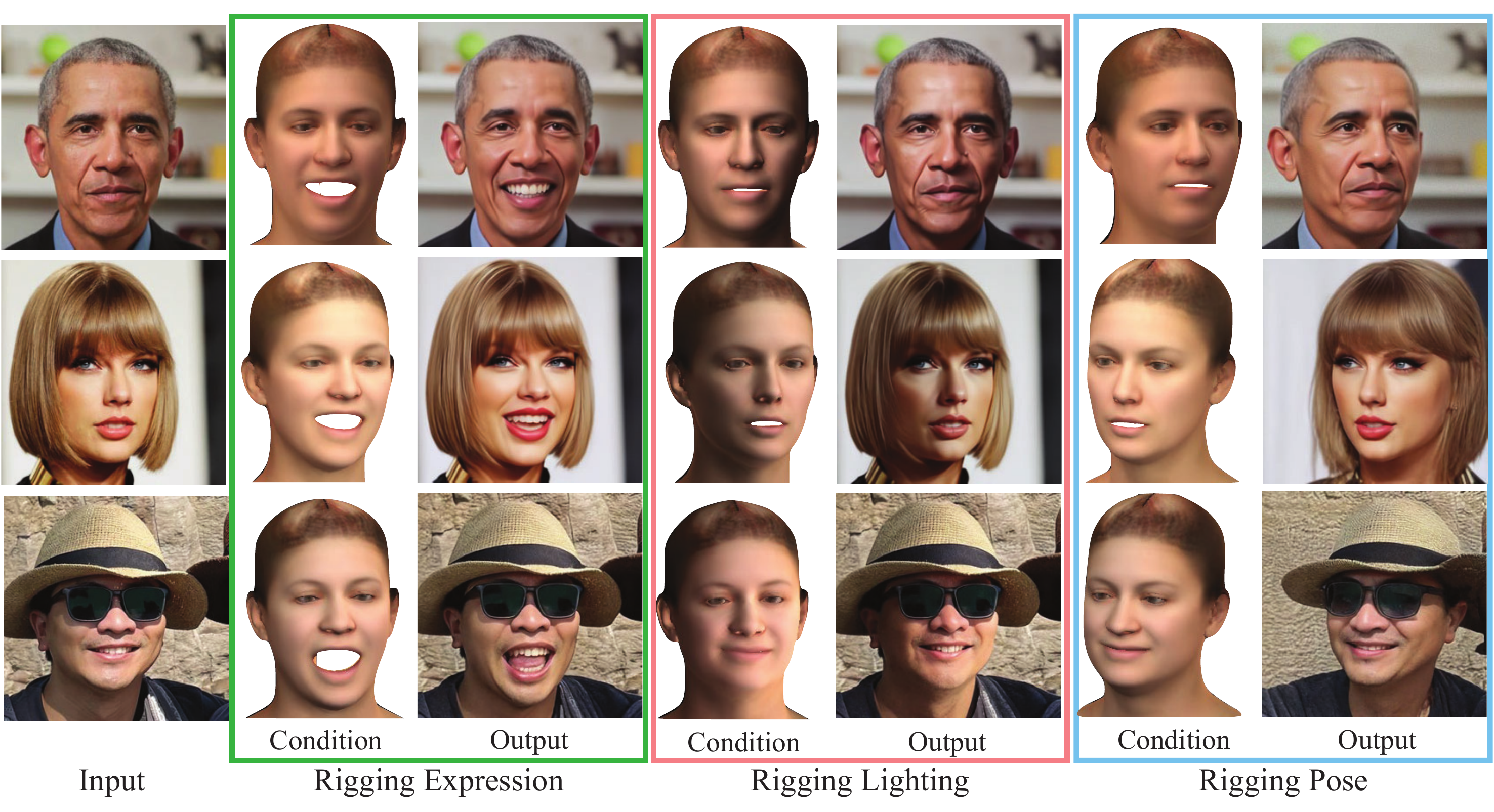}
    \vspace{-2em}
    \captionof{figure}{\model takes in coarse physical rendering as the condition to ``rig'' the input image with learned personal priors. The edited images respect the rendering conditions, preserve the identity, and exhibit high-frequency facial details.
    }
    \label{fig:teaser}
\end{center}%
}]

\blfootnote{$\dagger$ Work done during an internship at Adobe.}

\begin{abstract}
\vspace{-2mm}
We address the problem of learning person-specific facial priors from a small number (\eg, 20) of portrait photos of the same person.
This enables us to edit this specific person's facial appearance, such as expression and lighting, while preserving their identity and high-frequency facial details.
Key to our approach, which we dub \model, is a diffusion model conditioned on, or ``rigged by,'' crude 3D face models estimated from single in-the-wild images by an off-the-shelf estimator.
On a high level, \model learns to map simplistic renderings of 3D face models to realistic photos of a given person.
Specifically, \model is trained in two stages: It first learns generic facial priors from a large-scale face dataset and then person-specific priors from a small portrait photo collection of the  person of interest.
By learning the CGI-to-photo mapping with such personalized priors, \model can ``rig'' the lighting, facial expression, head pose, \etc of a portrait photo, conditioned only on coarse 3D models while preserving this person's identity and other high-frequency characteristics.
Qualitative and quantitative experiments show that \model outperforms existing approaches in both identity preservation and photorealism.
Please see the project website: \href{https://diffusionrig.github.io}{https://diffusionrig.github.io} for the supplemental material, video, code, and data.

\end{abstract}

\vspace{-5mm}
\section{Introduction}
\vspace{-1mm}

It is a longstanding problem in computer vision and graphics to photorealistically change the lighting, expression, head pose, \etc of a portrait photo while preserving the person's identity and high-frequency facial characteristics.
The difficulty of this problem stems from its fundamentally underconstrained nature, and prior work typically addresses this with zero-shot learning, where neural networks were trained on a large-scale dataset of different identities and tested on a new identity.
These methods ignore the fact that such generic facial priors often fail to capture the test identity's high-frequency facial characteristics, and multiple photos of the same person are often readily available in the person's personal photo albums, \eg, on a mobile phone.
In this work, we demonstrate that one can convincingly edit a person's facial appearance, such as lighting, expression, and head pose, while preserving their identity and other high-frequency facial details.
Our key insight is that we can first learn generic facial priors from a large-scale face dataset~\cite{karras2019style} and then finetune these generic priors into personalized ones using around 20 photos capturing the test identity.

When it comes to facial appearance editing, the natural question is what representation one uses to change lighting, expression, head pose, hairstyle, accessories, \etc.
Off-the-shelf 3D face estimators such as DECA~\cite{feng2021learning} can already extract, from an in-the-wild image, a parametric 3D face model that comprises parameters for lighting (spherical harmonics), expression, and head pose.
However, directly rendering these physical properties back into images yields CGI-looking results, as shown in the output columns of Figure~\ref{fig:teaser}.
The reasons are at least three-fold:
(a) The 3D face shape estimated is coarse, with mismatched face contours and misses high-frequency geometric details,
(b) the assumptions on reflectance (Lambertian) and lighting (spherical harmonics) are restrictive and insufficient for reproducing the reality, and
(c) 3D morphable models (3DMMs) simply cannot model all appearance aspects including hairstyle and accessories.
Nonetheless, such 3DMMs provide us with a useful representation that is amenable to ``appearance rigging'' since we can modify the facial expression and head pose by simply changing the 3DMM parameters as well as lighting by varying the spherical harmonics (SH) coefficients.

On the other hand, diffusion models~\cite{ho2020denoising} have recently gained popularity as an alternative to Generative Adversarial Networks (GANs)~\cite{goodfellow2020generative} for image generation.
Diff-AE~\cite{preechakul2022diffusion} further shows that when trained on the autoencoding task, diffusion models can provide a latent space for appearance editing.
In addition, diffusion models are able to map pixel-aligned features (such as noise maps in the vanilla diffusion model) to photorealistic images.
Although Diff-AE is capable of interpolating from, \eg, smile to no smile, after semantic labels are used to find the direction to move towards, it is unable to perform edits that require 3D understanding and that cannot be expressed by simple binary semantic labels.
Such 3D edits, including relighting and head pose change, are the focus of our work.

To combine the best of both worlds, we propose \model, a model that allows us to edit or ``rig'' the appearance (such as lighting and head pose) of a 3DMM and then produce a photorealistic edited image conditioned on our 3D edits.
Specifically, \model first extracts rough physical properties from single portrait photos using an off-the-shelf method~\cite{feng2021learning}, performs desired 3D edits in the 3DMM space, and finally uses a diffusion model~\cite{ho2020denoising} to map the edited ``physical buffers'' (surface normals, albedo, and Lambertian rendering) to photorealistic images.
Since the edited images should preserve the identity and high-frequency facial characteristics, we first train \model on the CelebA dataset~\cite{liu2015faceattributes} to learn generic facial priors so that \model knows how to map surface normals and the Lambertian rendering to a photorealistic image.
Note that because the physical buffers are coarse and do not contain sufficient identity information, this ``Stage 1 model'' provides no guarantee for identity preservation.
At the second stage, we finetune \model on a tiny dataset of roughly 20 images of one person of interest, producing a person-specific diffusion model mapping physical buffers to photos of just this person.
As discussed, there are appearance aspects not modeled by the 3DMM, including but not limited to hairstyle and accessories.
To provide our model with this additional information, we add an encoder branch that encodes the input image into a global latent code (``global'' in contrast to physical buffers that are pixel-aligned with the output image and hence ``local'').
This code is chosen to be low-dimensional in the hope of capturing just the aspects \emph{not} modeled by the 3DMM, such as hairstyle and eyeglasses.

In summary, our contributions are:
\begin{itemize}[noitemsep,topsep=0pt,parsep=0pt,partopsep=0pt]
    \item A deep learning model for 3D facial appearance editing (that modifies lighting, facial expression, head pose, \etc) trained using just images with no 3D label,
    \item A method to drive portrait photo generation using diffusion models with 3D morphable face models, and
    \item A two-stage training strategy that learns personalized facial priors on top of generic face priors, enabling editing that preserves identity and high-frequency details.
\end{itemize}

\begin{figure*}[ht]
    \centering
    \def\imW{1\linewidth}
        \includegraphics[width=\imW]{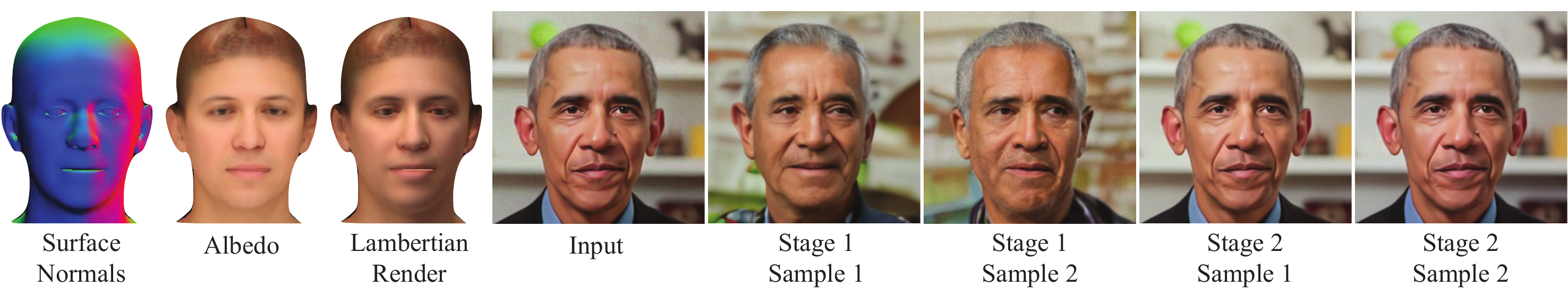}
    \vspace{-2em}
    \caption{\textbf{Reconstruction with \vs without personalized priors.}
    Given the input image and its conditions (surface normals, albedo, and Lambertian rendering) automatically extracted using DECA, Stage 1 learns only generic face priors and fails to reconstruct the identity in both of the randomly sampled reconstructions.
    With Stage 2, \model is able to faithfully reconstruct the input image using either of the two stochastically sampled noise maps.
    } 
    \label{fig:reconstruction}
    \vspace{-4mm}
\end{figure*}

\vspace{-2mm}
\section{Related Work}
\vspace{-1mm}

Our work is related to generative models, 3D Morphable Face Models (3DMMs), and personalized priors.

\vspace{-0.8em}
\myparagraph{Generative Modeling}
Since the proposal of early Generative Adversarial Networks (GANs) \cite{goodfellow2020generative}, researchers have made significant progress in generating photorealistic images of constrained classes, such as faces \cite{Karras_Laine_Aila_2019, Karras_Laine_Aittala_Hellsten_Lehtinen_Aila_2020,karras2021alias}.
Recently, denoising diffusion models \cite{Ho_Jain_Abbeel_2020}, which learn to denoise random noise images into photorealistic images, have shown impressive synthesis results and gained popularity as an alternative to GANs.
Different diffusion models are invented for faster sampling \cite{song2020denoising} (used in this work), conditional generation \cite{nichol2021improved,dhariwal2021diffusion}, and later pixel-aligned conditional generation \cite{saharia2022image}.
Similarly, we use pixel-aligned conditions, specifically surface normal, albedo, and Lambertian rendering images, as the condition that our diffusion model should satisfy.
Closely related to \model are Diffusion Autoencoders (Diff-AE) that learn a latent space of facial attributes (\eg, $+$smiling \vs $-$smiling) via the autoencoding task \cite{Preechakul_Chatthee_Wizadwongsa_Suwajanakorn_2022}.
Given binary labels of a certain attribute, the authors find the direction, along which the latent code should be pushed, to manipulate that attribute.
3D-aware generative models are a recent popular trend to combine 3D controllability with 2D image generation \cite{ghosh2020gif,tewari2020stylerig,zhao2022generative,tewari2022disentangled3d,Chan_Lin_Chan_Nagano_Pan_DeMello_Gallo_Guibas_Tremblay_Khamis_etal_2022,chan2021pi,Hao_Mallya_Belongie_Liu_2021,Tan_Fanello_Meka_Orts-Escolano_Tang_Pandey_Taylor_Tan_Zhang_2022}.

\vspace{-3mm}
\myparagraph{Facial Appearance Modeling}
3D Morphable Face Models or 3DMMs provide a valuable parameter space to describe (and in turn solve for) 3D facial characteristics \cite{blanz1999morphable}.
The FLAME face model learned from 4D scans is a widely-used 3DMM that supports shape, pose, and expression change \cite{li2017learning}.
We refer the reader to a recent survey paper on Morphable Face Models \cite{Egger_Smith_Tewari_Wuhrer_Zollhoefer_Beeler_Bernard_Bolkart_Kortylewski_Romdhani_etal_2020}.

RingNet regresses FLAME parameters from 2D images \cite{Sanyal_Bolkart_Feng_Black_2019}.
Also a learning-based method, DECA additionally predicts albedo and lighting in spherical harmonics (SH) from a single face image \cite{feng2021learning}.
An alternative to using 3DMMs for ``face de-rendering'' is directly predicting surface normals, albedo, and lighting in the image space, as in SfSNet \cite{Sengupta_Kanazawa_Castillo_Jacobs_2018}.
Although such approaches enjoy the benefit of being able to represent hair, accessories, \etc, image-space representations do not provide a physically meaningful parameter space for rigging like 3DMMs do.

The geometry, albedo, and lighting from 3DMM are still extremely coarse and far from reality.
The community has bridged the realism gap between 3DMM rendering and real photos through expensive hardware setups to capture fine-grained facial geometry \cite{Wang_Chen_Yu_Ma_Li_Liu_2022,Wuu_Zheng_Ardisson_Bali_Belko_Brockmeyer_Evans_Godisart_Ha_Hypes_etal_2022} and reflectance fields \cite{debevec2000acquiring}.
Neural network-based, implicit appearance models have also been proposed to address the infeasibility of explicitly describing the appearance with precise reflectance and lighting \cite{Lombardi_Saragih_Simon_Sheikh_2018,Bi_Lombardi_Saito_Simon_Wei_Mcphail_Ramamoorthi_Sheikh_Saragih,meka2019deep,remelli2022drivable,grassal2022neural,Shysheya_Zakharov_Aliev_Bashirov_Burkov_Iskakov_Ivakhnenko_Malkov_Pasechnik_Ulyanov_etal,R_Tewari_Dib_Weyrich_Bickel_Seidel_Pfister_Matusik_Chevallier_Elgharib_etal_2021,Zehni_Ghosh_Sridhar_Raman_2021,nestmeyer2020learning,sun2019single,zhang2020portrait}.

\vspace{-3mm}
\myparagraph{Personalized Priors}
Learning personal priors has been more widely discussed in super-resolution, face restoration, and inpainting, by using examplar imagery~\cite{wang2020multiple}, personal supplemental attributes~\cite{yu2018super}, an attention module with identity penalty~\cite{wang2022restoreformer}, or facial component dictionaries~\cite{li2020blind}.
Conditional portrait image editing also shares the objective of preserving the input identity~\cite{liu20223d, tewari2020pie}.

However, it remains a challenge how to compute an unbiased identity score, and these approaches do not explicitly learn personalized priors.

Closer to \model that learns a personal prior from a set of personal album of the person, MyStyle~\cite{nitzan2022mystyle} is a method to finetune a pre-trained StyleGAN model to achieve a generative model for a specific identity, while preserving the expressiveness of the latent space.
However, it does not support precise 3D rigging to control the generation and requires a much larger personal dataset to obtain a smooth personalized latent space.
\model, on the other hand, focuses on controllable image editing and achieves the smooth editing naturally with the continuous physical space as conditions.

\vspace{-1mm}
\section{Preliminaries}

We provide the background knowledge for the two building blocks---3D Morphable Face Models (3DMMs) and Denoising Diffusion Probabilistic Models (DDPMs).

\begin{figure*}[t!]
\centering
\includegraphics[width=1.0\textwidth]{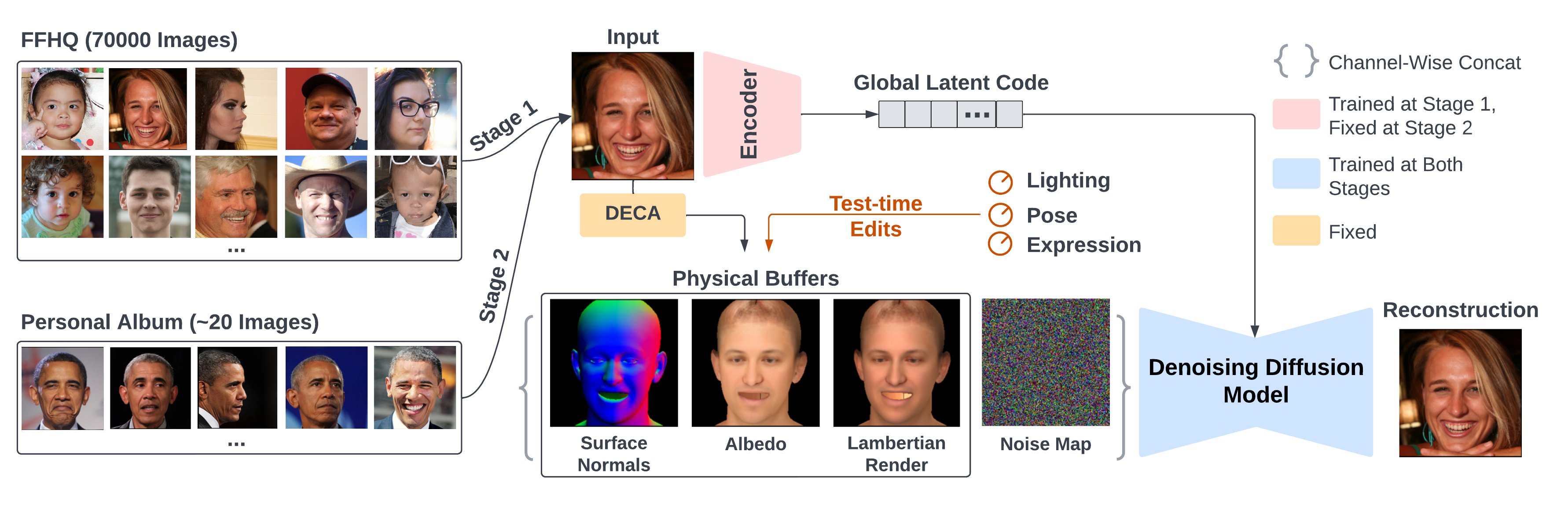}
\vspace{-2.9em}
\caption{
\textbf{\model overview.}
The input to our model is a set of physical buffers reconstructed from the input, a random noise map, and a global latent code that encodes nuance features not modeled by the physical buffers.
At Stage 1, we train our model on a large face dataset to learn generic face priors.
At Stage 2, we keep the global latent code encoder frozen and fine-tune the diffusion model to learn personalized priors.
}
\vspace{-1em}
\label{fig:pipeline}
\end{figure*}

\vspace{-1mm}
\subsection{3D Morphable Face Models}

3D Morphable Face Models or 3DMMs are parametric models that use a compact latent space (handcrafted or learned from scans) to represent the head pose, face geometry, facial expression, \etc \cite{blanz1999morphable,li2017learning}.
In this paper, we employ FLAME \cite{li2017learning}, a popular 3DMM using standard vertex-based linear blend skinning with corrective blendshapes and representing a face mesh with pose, shape, and expression parameters.
Although FLAME provides a compact and physically meaningful space for face \emph{geometry}, it does not provide descriptions for \emph{appearance}.
To this end, DECA \cite{feng2021learning} uses FLAME and additionally models facial appearance with Lambertian reflectance and Spherical Harmonics (SH) lighting. 
Trained on a large dataset, DECA predicts albedo, SH lighting, and the FLAME parameters from a single portrait image. 
We utilize DECA to generate \emph{rough} 3D representations that support easy ``rigging'' by editing the FLAME parameters, albedo, and/or SH lighting.
As we show in Figure~\ref{fig:reconstruction}, the realism gap between DECA rendering and real photos is significant, calling for measurements post-editing.

\subsection{Denoising Diffusion Probabilistic Models}

Denoising Diffusion Probabilistic Models (DDPMs) \cite{ho2020denoising, nichol2021improved, dhariwal2021diffusion} are a class of generative models that take random noise images as input and denoise the images progressively to produce photorealistic images.
This generation process can be seen as the reverse of the diffusion process that gradually adds noise to images.
The key component of DDPMs is a denoising network $f_{\theta}$.
During training, it takes a noisy image $x_t$ and a timestep $t$ ($1 \leq t\leq T $), and predicts the noise at time $t$: $\epsilon_t$.
More formally, the predicted noise at time $t$ is $\hat{\epsilon_t} = f_\theta(x_t, t)$, where $x_t = \alpha_t x_0 + \sqrt{1 - \alpha_t^2}\epsilon_t$, $\epsilon_t$ is a random, normally distributed noise image, and $\alpha_t$ is a hyperparamter that gradually increases noise level of $x_t$ with each step of the forward process.
The loss is computed on the distance between $\epsilon_t$ and $\hat{\epsilon_t}$.
Therefore, the trained model can generate images by taking as input a random noise image and gradually denoising it to a photorealistic one. 

\vspace{-0.5em}
\section{Method}

To enable personalized appearance editing, our model, which we dub \model, needs to (a) generate images based on different appearance conditions, such as novel lighting, and (b) learn personal priors so that the person's identity is not altered during editing.

To this end, we design a two-stage training pipeline as shown in Figure \ref{fig:pipeline}. 
At the first stage, the model learns generic face priors by being trained to reconstruct portrait images given their underlying ``appearance conditions'' represented as physical buffers automatically extracted using an off-the-shelf estimator.
At the second stage, we finetune our model using portrait photos of just one person so that the model learns personalized priors, which are necessary to prevent identity shift during appearance editing.

\subsection{Learning Generic Face Priors}

Our first stage is designed to learn facial priors that enable photorealistic image synthesis conditioned on physical constraints like lighting.
For the physical conditioning, we use DECA \cite{feng2021learning} to produce the physical parameters including the FLAME \cite{li2017learning} parameters (shape $\boldsymbol{\beta}$, expression $\boldsymbol{\psi}$, and pose $\boldsymbol{\theta}$), albedo $\boldsymbol{\alpha}$, (orthographic) camera $\textbf{c}$, and (spherical harmonics) lighting $\textbf{l}$ from the input portrait image.
We then use the Lambertian reflectance to render these physical properties into three buffers: surface normals, albedo, and Lambertian rendering.
Although these physical buffers provide pixel-aligned descriptions of the facial geometry, albedo, and lighting, they are rather coarse and nowhere close to photorealistic images (see the Lambertian rendering in Figures \ref{fig:teaser} and \ref{fig:reconstruction}).
Still, using these buffers, we can ``rig'' our generative model in a disentangled, physically meaningful way by changing the DECA parameters.
For photorealistic image synthesis, we use a Denoising Diffusion Probabilistic Model (DDPM) as our generator because DDPMs can naturally take pixel-aligned conditions (more advantageous than latent code conditions as shown in Section \ref{sec:ablation}) to drive the generation process.

Besides the pixel-aligned physical buffers, we keep the random noise images in DDPMs to explain the stochasticity during generation.
In addition to the pixel-aligned buffers and noise map, we need another condition to encode \emph{global} appearance information (as opposed to local information such as local surface normals) that is not modeled by the physical buffers, such as hair, hat,  glasses, and the image background.
Therefore, our diffusion model takes both physical buffers and a learned global latent code as conditions for image synthesis.
Formally, our model can be described as $\hat{\epsilon_t} = f_{\theta}\left([x_t, z], t, \phi_\theta\left(x_0\right)\right)$
where $x_t$ is the noisy image at timestep $t$, $z$ represent the physical buffers, $x_0$ is the original image, $\hat{\epsilon_t}$ is the predicted noise, and $f_\theta$ and $\phi_\theta$ are the denoising model and the global latent encoder, respectively.

It is theoretically possible that the global latent code also encodes local geometry, albedo, and/or illumination information, which could lead to the diffusion model ignoring the physical buffers entirely.
Empirically, we find that the network learns to use the physical buffers for local information and does not rely on the global latent code, possibly because these buffers are pixel-aligned with the ground truth and thus more easily leveraged by the model.

\subsection{Learning Personalized Priors}
\vspace{-1mm}
After learning the generic facial priors at the first stage, \model is able to generate photorealistic images given coarse physical buffers.
The next step is to learn personalized priors for a given person to avoid identity shift during appearance editing.
Personal priors are crucial to preserving identity and high-frequency facial characteristics, as shown in Figure \ref{fig:reconstruction}.
We achieve this by finetuning our denoising model on a specific person's photo album of around 20 images.
During the finetuning stage, the denoising model learns the person's identity information.
We fix the global encoder from the previous stage since it has learned to encode global image information not modeled by the physical buffers (which we want to retain).
We show that this approach is simple and yet effective compared with GANs that need careful tuning, as mentioned in MyStyle \cite{nitzan2022mystyle}.

For this small personalized dataset, we also extract the DECA parameters first.
However, since DECA is a single-image estimator, its output is sensitive to extreme poses or expressions.
Under the assumption that the general shape of a person's face does not change drastically within a reasonable period of time, we compute the mean of the shape parameters in FLAME over all the images in the album and use that mean shape when conditioning \model. 

\subsection{Model Architecture}
\vspace{-1mm}
\model consists of two trainable parts: a denoising model $f_{\theta}$ and a global encoder $\phi_\theta$.
The architecture of our denoising model is based on ADM \cite{dhariwal2021diffusion} with modifications to reduce computational cost and take an additional global latent code as input.
For the global code, we use the same method that ADM uses for their time embedding: We scale and shift the features in each layer using the global latent code. 
The encoder is simply a ResNet-18 \cite{he2016deep} and we use the output features as the global latent codes.

Our loss function is a P2 weight loss \cite{choi2022perception} that computes distances between predicted and ground-truth noises:
$\mathcal{L} = \lambda'_t ||\hat{\epsilon_t} - \epsilon_t||_2^2$,
where $\lambda'_t$ is a hyperparameter to control the loss weight at different timesteps.
We empirically find that the P2 weight loss speeds up the training process and generates high-quality images compared with a constant loss weight.

\subsection{Implementation Details}
\vspace{-1mm}
During the first stage, we train \model on the FFHQ dataset \cite{karras2019style}, which contains 70,000 images.
With Adam \cite{kingma2014adam} as the optimizer with a learning rate of $10^{-4}$, we train \model for 50,000 iterations with a batch size of 256 (so the total number of samples seen by the model is 12,800,000).
During the second stage, we use only 10--20 images of a single person.
In the following, we show results for four celebrities (Obama, Biden, Swift, and Harris) and two non-celebrities.
Please see the supplemental material and video for more results including more identities.
We use 20 images for each person except for Harris, for whom we use only 10, and for the ablation study on the number of training images. We provide the personal photo album of two identities in the supplemental material.
We finetune our model on each small dataset for 5,000 iterations with a batch size of 4 (so the total number of seen samples during finetuning is 20,000).
We furthermore decrease the learning rate to $10^{-5}$ for the second stage.
Training for the first stage takes around 15 hours using eight A100 GPUs, and the Stage 2 finetuning completes within 30 minutes on a single V100 GPU.

\begin{figure*}[ht]
    \centering
    \setlength{\tabcolsep}{1pt}
    \def\imW{0.1\linewidth}
    \begin{tabular}{cccc@{\hskip 6pt}ccc@{\hskip 6pt}ccc}

        \rotatebox[origin=c]{90}{Original} &
        \raisebox{-.5\height}{\includegraphics[width=\imW]{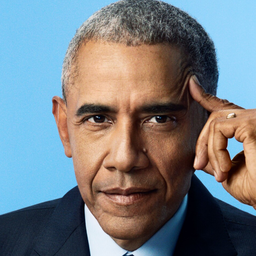}} &
        \raisebox{-.5\height}{\includegraphics[width=\imW]{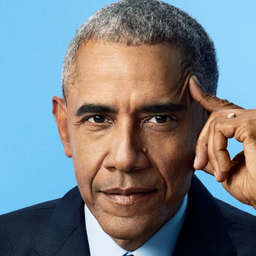}} &
        \raisebox{-.5\height}{\includegraphics[width=\imW]{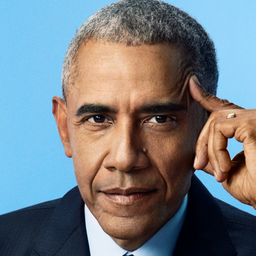}} &
        
        \raisebox{-.5\height}{\includegraphics[width=\imW]{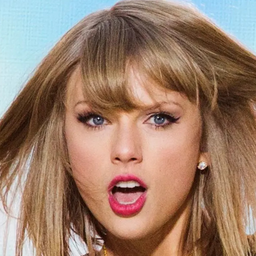}} &
        \raisebox{-.5\height}{\includegraphics[width=\imW]{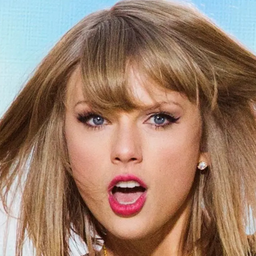}} &
        \raisebox{-.5\height}{\includegraphics[width=\imW]{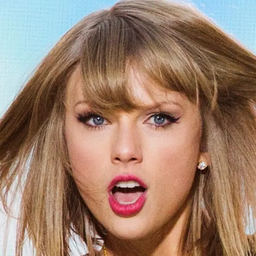}} &
        
        \raisebox{-.5\height}{\includegraphics[width=\imW]{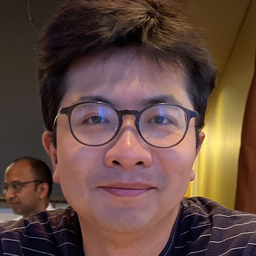}} &
        \raisebox{-.5\height}{\includegraphics[width=\imW]{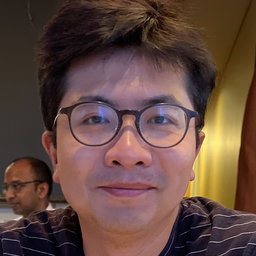}} &
        \raisebox{-.5\height}{\includegraphics[width=\imW]{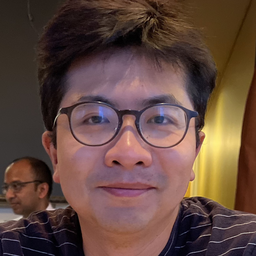}} \vspace{3pt}\\
        
        \rotatebox[origin=c]{90}{DECA} &
        \raisebox{-.5\height}{\includegraphics[width=\imW]{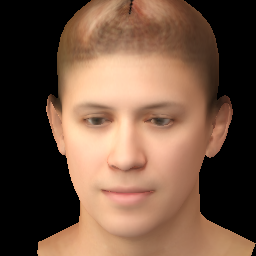}} &
        \raisebox{-.5\height}{\includegraphics[width=\imW]{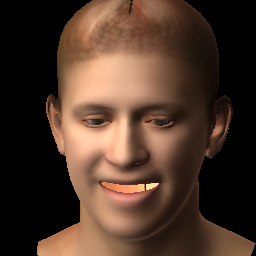}} &
        \raisebox{-.5\height}{\includegraphics[width=\imW]{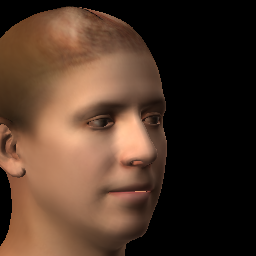}} &
        
        \raisebox{-.5\height}{\includegraphics[width=\imW]{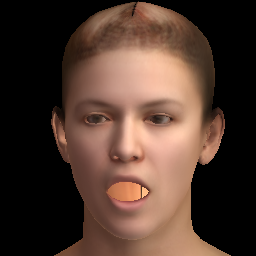}} &
        \raisebox{-.5\height}{\includegraphics[width=\imW]{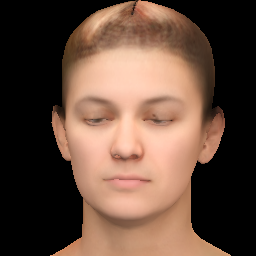}} &
        \raisebox{-.5\height}{\includegraphics[width=\imW]{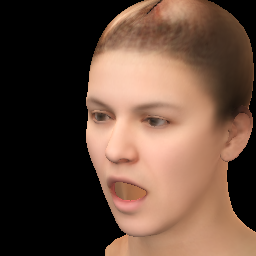}} &
        
        \raisebox{-.5\height}{\includegraphics[width=\imW]{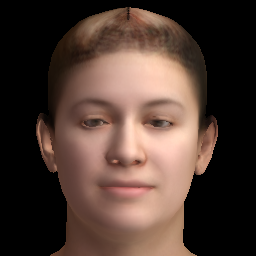}} &
        \raisebox{-.5\height}{\includegraphics[width=\imW]{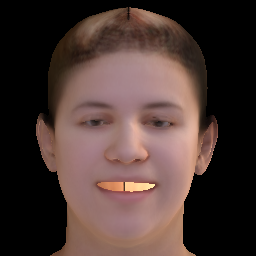}} &
        \raisebox{-.5\height}{\includegraphics[width=\imW]{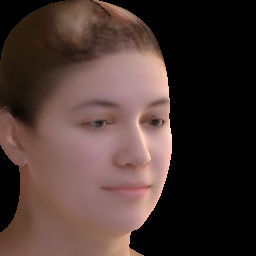}} \vspace{3pt} \\

        \rotatebox[origin=c]{90}{HeadNerf} &
        \raisebox{-.5\height}{\includegraphics[width=\imW]{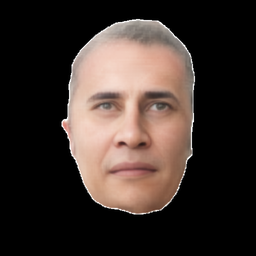}} &
        \raisebox{-.5\height}{\includegraphics[width=\imW]{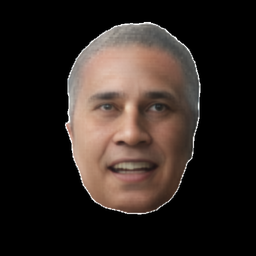}} &
        \raisebox{-.5\height}{\includegraphics[width=\imW]{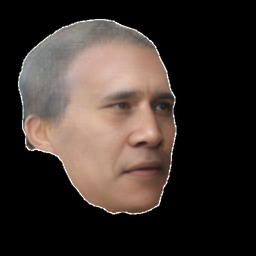}} &
        
        \raisebox{-.5\height}{\includegraphics[width=\imW]{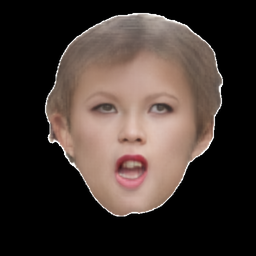}} &
        \raisebox{-.5\height}{\includegraphics[width=\imW]{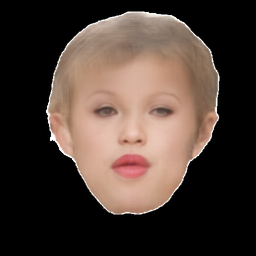}} &
        \raisebox{-.5\height}{\includegraphics[width=\imW]{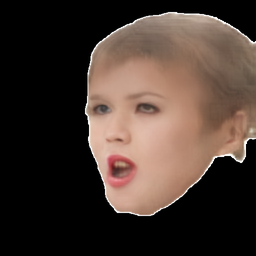}} &
        
        \raisebox{-.5\height}{\includegraphics[width=\imW]{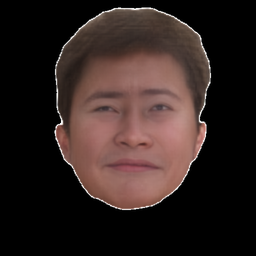}} &
        \raisebox{-.5\height}{\includegraphics[width=\imW]{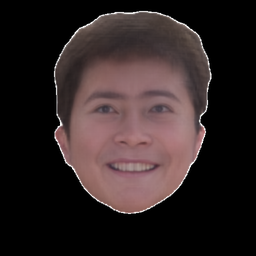}} &
        \raisebox{-.5\height}{\includegraphics[width=\imW]{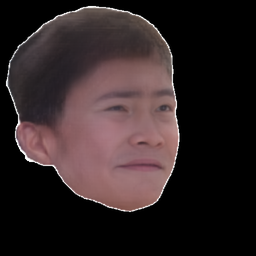}} \vspace{3pt} \\
        
        \rotatebox[origin=c]{90}{GIF} &
        \raisebox{-.5\height}{\includegraphics[width=\imW]{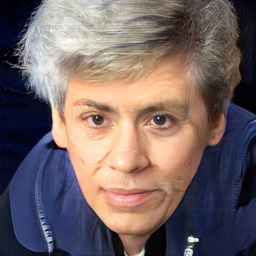}} &
        \raisebox{-.5\height}{\includegraphics[width=\imW]{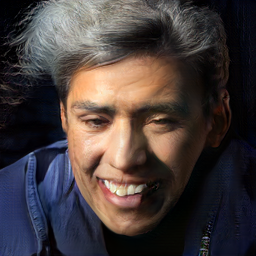}} &
        \raisebox{-.5\height}{\includegraphics[width=\imW]{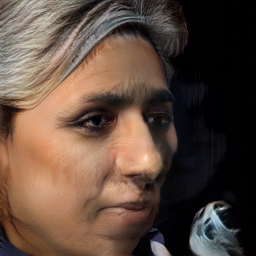}} &
        
        \raisebox{-.5\height}{\includegraphics[width=\imW]{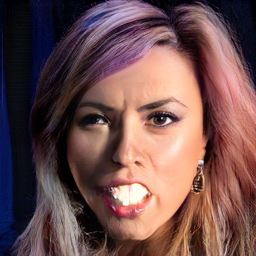}} &
        \raisebox{-.5\height}{\includegraphics[width=\imW]{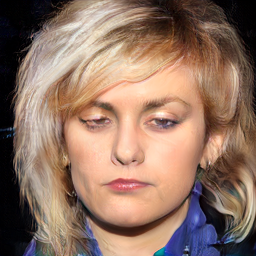}} &
        \raisebox{-.5\height}{\includegraphics[width=\imW]{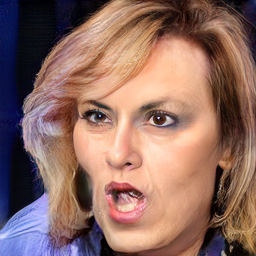}} &
        
        \raisebox{-.5\height}{\includegraphics[width=\imW]{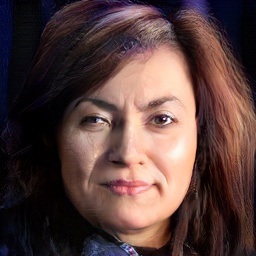}} &
        \raisebox{-.5\height}{\includegraphics[width=\imW]{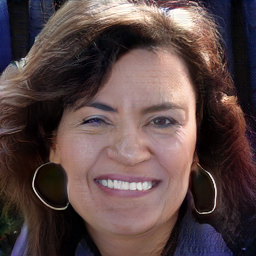}} &
        \raisebox{-.5\height}{\includegraphics[width=\imW]{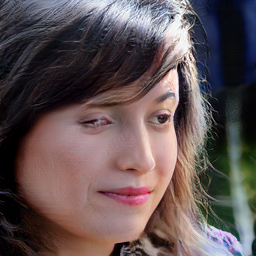}} \vspace{3pt} \\
        
        \rotatebox[origin=c]{90}{MyStyle} &
        \raisebox{-.5\height}{\includegraphics[width=\imW]{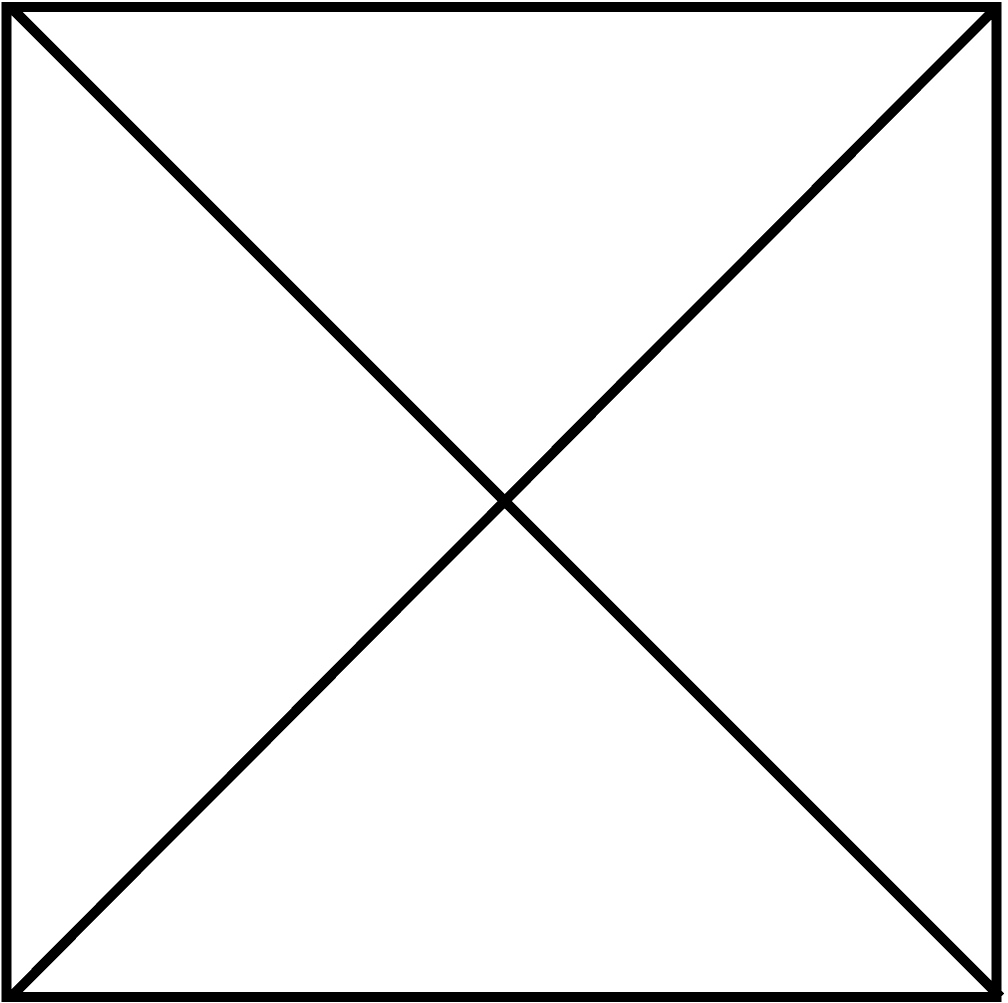}} &
        \raisebox{-.5\height}{\includegraphics[width=\imW]{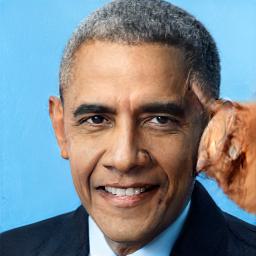}} &
        \raisebox{-.5\height}{\includegraphics[width=\imW]{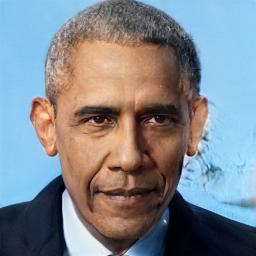}} &
        
        \raisebox{-.5\height}{\includegraphics[width=\imW]{figures/relighting/not_supported.png}} &
        \raisebox{-.5\height}{\includegraphics[width=\imW]{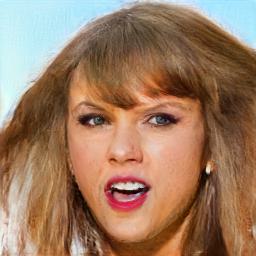}} &
        \raisebox{-.5\height}{\includegraphics[width=\imW]{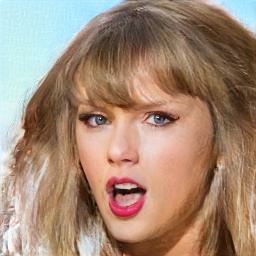}} &
        
        \raisebox{-.5\height}{\includegraphics[width=\imW]{figures/relighting/not_supported.png}} &
        \raisebox{-.5\height}{\includegraphics[width=\imW]{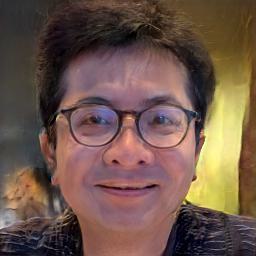}} &
        \raisebox{-.5\height}{\includegraphics[width=\imW]{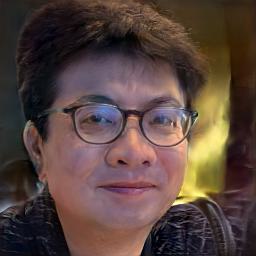}}\vspace{3pt}  \\
        
        \rotatebox[origin=c]{90}{Ours} &
        \raisebox{-.5\height}{\includegraphics[width=\imW]{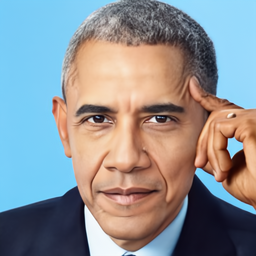}} &
        \raisebox{-.5\height}{\includegraphics[width=\imW]{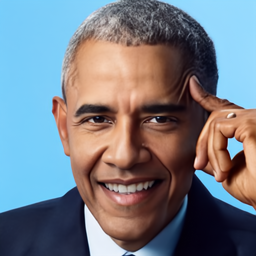}} &
        \raisebox{-.5\height}{\includegraphics[width=\imW]{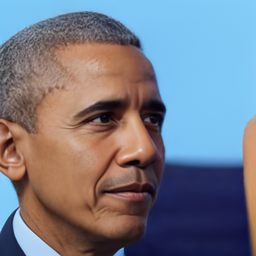}} &
        
        \raisebox{-.5\height}{\includegraphics[width=\imW]{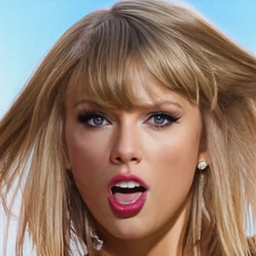}} &
        \raisebox{-.5\height}{\includegraphics[width=\imW]{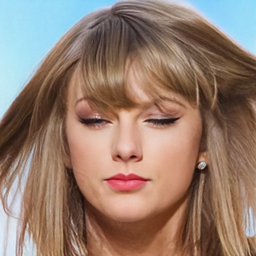}} &
        \raisebox{-.5\height}{\includegraphics[width=\imW]{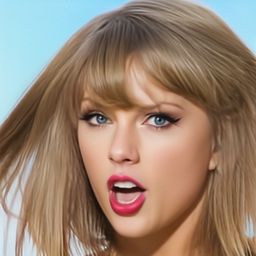}} &
        
        \raisebox{-.5\height}{\includegraphics[width=\imW]{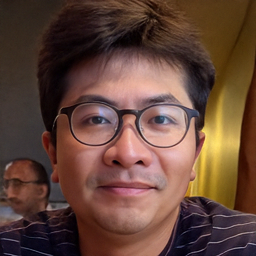}} &
        \raisebox{-.5\height}{\includegraphics[width=\imW]{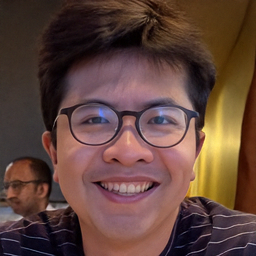}} &
        \raisebox{-.5\height}{\includegraphics[width=\imW]{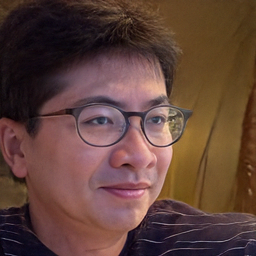}} \vspace{3pt} \\
        
         & Lighting & Expression & Pose & Lighting & Expression & Pose & Lighting & Expression & Pose \\

    \end{tabular}
    \vspace{-7pt}
    \caption{\textbf{Appearance editing.}
    \model achieves convincing appearance edits while preserving the individual's identity using only 20 images per identity.
    GIF creates realistic-looking images but does not use personalized priors, leading to significant identity shifts.
    MyStyle is unable to make dramatic changes to the expression or pose without artifacts or minor identity shifts.
    In addition, MyStyle does not trivially support controllable relighting, so the corresponding fields have been left empty.}
    \label{fig:rig_results}
    \vspace{-5mm}
\end{figure*}

\vspace{-2mm}
\section{Experiments}

We first show how to edit a person's appearance (\eg, facial expression, lighting, and head pose) by modifying the physical buffers that condition the model.
We then demonstrate how to rig, with the global latent code, other aspects of a person's appearance not modeled by the physical buffers such as hairstyle and accessories.
By swapping in the global latent code from another image, we can transfer portrait characteristics, such as hairstyle, accessories including glasses, and/or the image background, while preserving the physical properties (\eg, identity, pose, expression, and lighting) from the original image.
Finally, we show the power of the learned personal priors by conditioning, for example, an Obama model on both the physical buffers and global latent code from a different person (to ``Obama-fy'' that person).
\vspace{-1mm}
\subsection{Rigging Appearance With Physical Buffers}
\vspace{-1mm}
In this section, we use our personalized model to rig the appearance with physical buffers.
We show three different types of appearance rigging: relighting, expression change, and pose change.
For relighting, we use different Spherical Harmonics (SH) parameters for producing the Lambertian rendering. 
To change the expression, we modify the expression and jaw rotation parameters of FLAME (the last three parameters of the pose vector).
To vary the pose, we modify the head rotation parameters (the first three parameters of the pose vector).
The 64-dimensional global latent code is produced by encoding the input image and remains unchanged when editing appearance.

Our results are displayed in Figure \ref{fig:rig_results}, where we depict three identities: two celebrities and one daily user. All the images have a resolution of 256$\times$256. Additionally, 512$\times$512 results can be found in the supplemental material.
We compare our method against DECA \cite{feng2021learning}, HeadNerf \cite{hong2021headnerf}, GIF \cite{ghosh2020gif}, and MyStyle \cite{nitzan2022mystyle}, of which the first two are 3D face model estimation methods, and the latter two are GAN-based approaches.
As Figure~\ref{fig:rig_results} shows, while GIF is capable of rigging the appearance by changing the expression and pose, it fails to preserve the individual's identity.
\model and MyStyle, on the other hand, are both personalized models that are able to preserve the identity.
However, since our method is directly conditioned on physical buffers, we can rig the appearance in a physically-based manner, whereas MyStyle needs to search for and step into a certain direction within the latent space to produce the target appearance, limiting its controllability, interpretability, and capacity for dramatic appearance changes.
We also observe more artifacts for MyStyle when doing appearance editing, which is likely due to the use of too few images during finetuning the StyleGAN model.

\subsection{Rigging Appearance With Global Latent Code}
\vspace{-1mm}
By design, \model finds it easier to learn what physical buffers can describe from the pixel-aligned buffers than from the global latent code.
The latent code thus encodes what physical buffers cannot describe including background, makeup, and hairstyle.
In this part, we change the global latent code to show its effects on the generated images.

\begin{figure}[h!]
    \centering
    \setlength{\tabcolsep}{2pt}
    \begin{tabular}{ccccc}
    \includegraphics[width=1.0\linewidth]{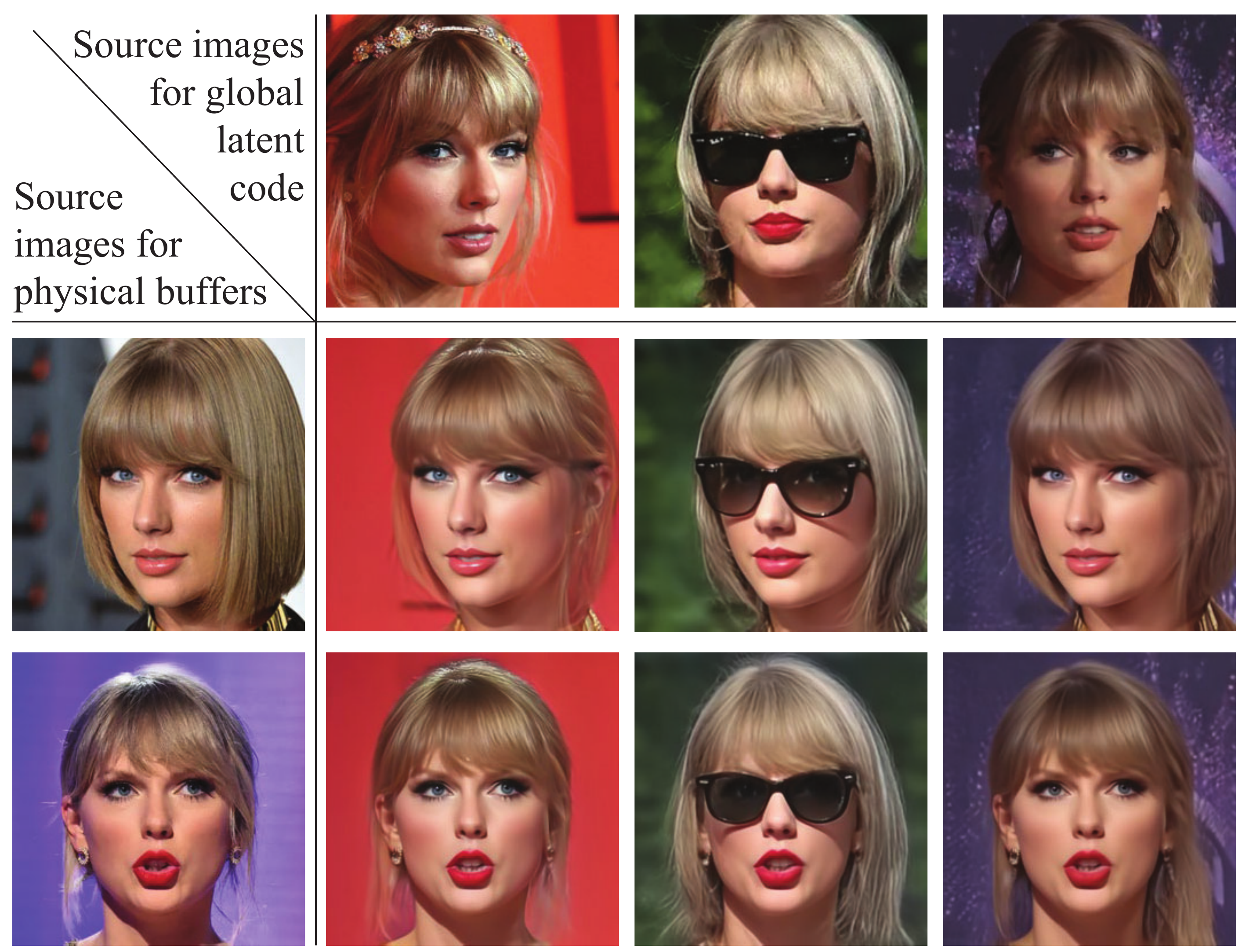}

    \end{tabular}
    \vspace{-10pt}
    \caption{
    \textbf{Mix and match of physical buffers and global latent code.}
    We mix the physical buffers from one image and the global latent code from another image to demonstrate how the two conditions encode disentangled information.}
    \label{fig:globallatent}
    \vspace{-1em}
\end{figure}

In Figure \ref{fig:globallatent}, we show a 2$\times$3 matrix of generated images.
Along the horizontal axis, we swap in the global latent code from another image of the same person while keeping the physical buffers identical (\ie, same physical buffers but different global codes).
Along the vertical axis, we replace the physical buffers while keeping the same global latent code (\ie, same global code but different physical buffers).
We can see that geometry information, such as head pose and expression, is preserved for each row, which shows that only the physical buffers (not the latent code) contain such information.
This means that in \model these physical properties are well disentangled from each other and from other appearance properties that physical buffers cannot describe.
On the other hand, the information hard to model explicitly, including image background, glasses, and hair style/color, is encoded in the global latent code. 

\vspace{-2mm}
\subsection{Identity Transfer With Learned Priors}
\begin{figure}[h!]
    \centering
    \newcommand{\cfbox}[2]{%
        \colorlet{currentcolor}{.}%
        {\color{#1}%
        \fbox{\color{currentcolor}#2}}%
    }
    \setlength{\tabcolsep}{1pt}
    \def\imW{0.23\linewidth}
    \setlength{\fboxsep}{0.5pt}%
    \setlength{\fboxrule}{1pt}%
    \begin{tabular}{ccccc}
        Obama-fy & Hopkins-fy & Swift-fy & Williams-fy 
        \\
        \cfbox{green}{\includegraphics[width=\imW]{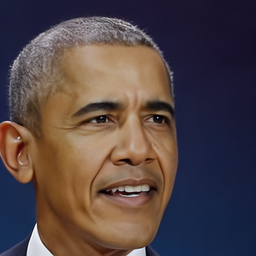}} & 
        \includegraphics[width=\imW]{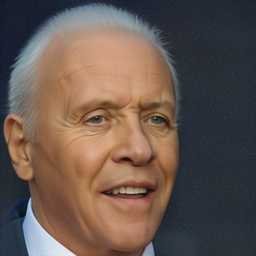} &
        \includegraphics[width=\imW]{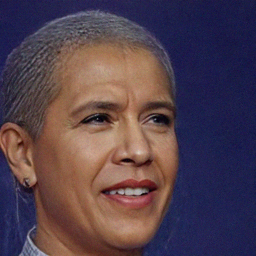}  &
        \includegraphics[width=\imW]{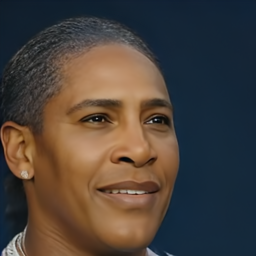} 
        \\
        \includegraphics[width=\imW]{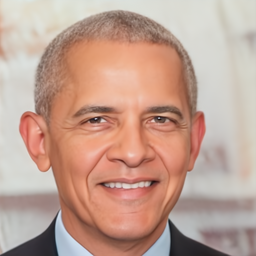} & 
        \cfbox{green}{\includegraphics[width=\imW]{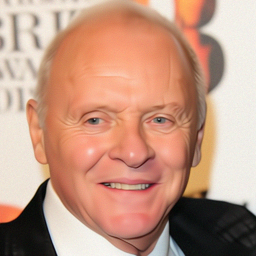}} &
        \includegraphics[width=\imW]{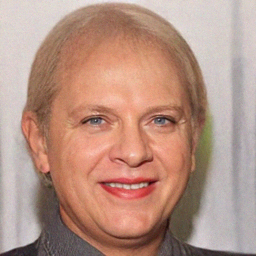}  &
        \includegraphics[width=\imW]{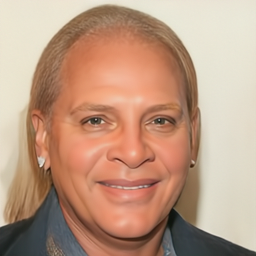}
        \\
        \includegraphics[width=\imW]{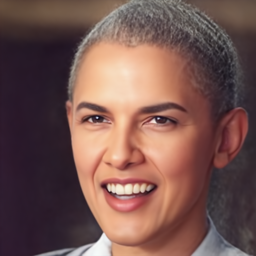} & 
        \includegraphics[width=\imW]{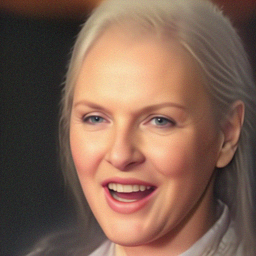} &
        \cfbox{green}{\includegraphics[width=\imW]{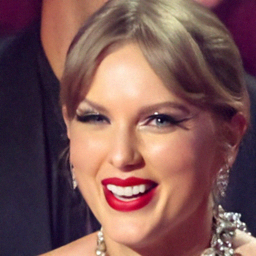}}  &
        \includegraphics[width=\imW]{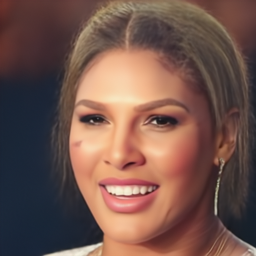}
        \\
        \includegraphics[width=\imW]{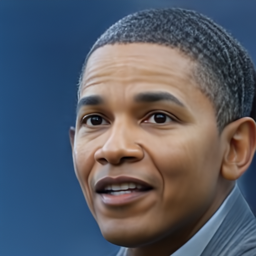} & 
        \includegraphics[width=\imW]{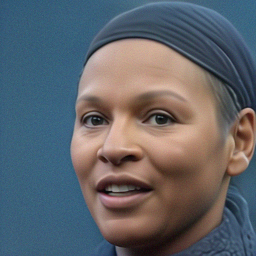} &
        \includegraphics[width=\imW]{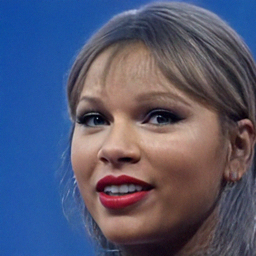}  &
        \cfbox{green}{\includegraphics[width=\imW]{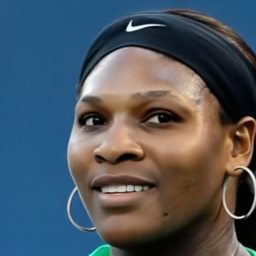}}
    \end{tabular}
    \vspace{-8pt}
    \caption{\textbf{Swapping personalized models}.
We demonstrate the power of personalized priors by running one person's model on other identities.
This creates the effect of ``adding'' one person's identity to another person.
The images with green borders are ``no-swap'' results where the corresponding person's model is used.
}

    \label{fig:xxfy}
\end{figure}

\vspace{-1mm}
In previous sections, we saw what information the physical buffers and the global latent code encode. Now, we demonstrate what information is encoded in the personalized diffusion models' weights.
Here, we keep both physical buffers and global latent code the same but exchange the personalized model itself with another person's personalized model (\ie, model swapping without code or buffer swapping).
The results of this experiment for four identities are shown in Figure \ref{fig:xxfy}.
Each row uses the same physical buffers and latent code but another personalized model.
Each column uses the same personalized model but different physical buffers and latent code.
For example, the column ``Obama-fy'' shows four images that are generated by Obama's personal model but using the other celebrities' images as input.
We see that across each row, while all inputs (physical buffers plus global latent code) are the same, the four different personalized models output different identities.
These results further corroborate that our model is able to learn personalized priors from a small dataset.

\subsection{Baseline Comparisons \& Evaluation Metrics}
\vspace{-1mm}
We evaluate our \model quantitatively in three aspects: rigging quality, identity preservation, and photorealism, since these three qualities are the most important for our personalized appearance editing.

\vspace{-3mm}
\myparagraph{DECA Re-Inference Error}
We follow the same setup as in GIF~\cite{ghosh2020gif} to compute the DECA re-inference error.
To evaluate relighting quality, we directly compute the RMSE on the re-inferred spherical harmonics. We show our results in Table \ref{tab:reinference}.
For our model, we also evaluate two ablated versions: ``vector cond.'' and ``feature cond.''
Instead of using pixel-aligned physical buffers as the condition, we use DECA's output parameters and features computed from physical buffers as conditions in our two ablated models.
More details can be found in Section \ref{para:ab_cond}.

\begin{table}[!h]
    \centering
    \scalebox{0.8}{
    \begin{tabular}{l|c|c|c|c}
        & Light $\downarrow$ & Shape $\downarrow$ & Exp.\ $\downarrow$ & Pose $\downarrow$ \\
        \hline 
        GIF \cite{ghosh2020gif} & 13.8& \textbf{3.0} & 5.0 & 5.6 \\
        GIF, vector cond.\ \cite{ghosh2020gif} & -- & 3.4 & 23.1 & 29.7 \\
        \hline
        \model (Ours) & \textbf{11.2} & 4.3 & \textbf{2.8} & \textbf{4.2} \\
        \model, vector cond. & 15.5 & 10.7 & 8.8 & 14.0  \\
        \model, feature cond. & 27.0 & 5.3 & 4.1 & 21.6 \\
    \end{tabular}}
    \vspace{-.5em}
    \caption{\textbf{RMSE of DECA re-inference}.
    All numbers are multiplies of $10^{-3}$.
    We generate 1,000 images for evaluation.
    For shape, expression, and pose, the RMSE is computed on rendered FLAME faces.
    For lighting, the RMSE is computed on re-inferred spherical harmonics directly.
    We only use our Stage 1 model since GIF is not a personalized model.
    Numbers for GIF and its vector-conditioned variant are cited from the original paper \cite{ghosh2020gif}.
    }
    \vspace{-1em}
    \label{tab:reinference}
\end{table}
\vspace{-2mm}
\myparagraph{Face Re-Identification Error}
An important metric for evaluating this work is whether \model can preserve the identity after appearance editing, since identity shift is a notorious problem in generative model-based editing.
To this end, we run a widely popular face re-identification network \cite{dlib09} to automatically determine if the edited and original images are of the same person.
As Table~\ref{tbl:re_id_and_user_study} shows, both MyStyle \cite{nitzan2022mystyle} and \model preserve the identity in all 400 expression-edited images of Obama and another 400 of Swift.
That said, for dramatic changes such as head pose change, \model preserves the identity better than MyStyle, as also demonstrated by Figure~\ref{fig:rig_results}.
One caveat of this error metric, though, is the obvious degenerate solution of not applying any edit at all, thereby achieving a perfect score.
We refer the reader to Figure~\ref{fig:rig_results} and Table~\ref{tab:reinference}, which show that \model avoids this degenerate solution.
\vspace{-3mm}
\myparagraph{User Study}

To further evaluate both the photorealism and identity preservation of images from \model against MyStyle, we conduct a user study involving Amazon Mechanical Turk.
During the study, we show pairs of images, where the left image is an original image from the real image dataset, and the right image is a generated one.
We occasionally include some real images on the right, too, for consistency check and quality control.
We then ask the users whether the right image is a real image of the person on the left (so both photorealism and identity preservation are probed).
We generate images that include either an expression or pose change for both \model and MyStyle.
We report our results in Table \ref{tbl:re_id_and_user_study}. 

\begin{table}[]
    \centering
    \scalebox{0.75}{
    \begin{tabular}{l|c|c|c|c|c|c|c|c}
      & \multicolumn{4}{c|}{Auto.\ Face Re-ID $\uparrow$} & \multicolumn{4}{c}{User Study $\uparrow$}\\
     \hline
      & \multicolumn{4}{c|}{Obama and Swift} & \multicolumn{2}{c|}{Obama} & \multicolumn{2}{c}{Swift} \\
         & \multicolumn{2}{c|}{Expr.} & \multicolumn{2}{c|}{Pose} & Expr.\ & Pose & Expr.\ & Pose \\
        MyStyle & \multicolumn{2}{c|}{\textbf{100\%}} & \multicolumn{2}{c|}{97.9\%} & 79.4\% & 78.0\% & 64.5\% & 62.5\%\\
        \hline
        Ours & \multicolumn{2}{c|}{\textbf{100\%}} & \multicolumn{2}{c|}{\textbf{99.3\%}} & \textbf{87.2\%} & \textbf{86.5\%} & \textbf{82.4\%} & \textbf{80.2\%} \\

    \end{tabular}}
    \vspace{-1.5mm}
    \caption{
    \textbf{\model \vs MyStyle \cite{nitzan2022mystyle}} in expression and pose editing, as measured by an automatic face re-ID error \cite{dlib09} (which has an obvious flaw; see text) as well as a user study on both realism and identity preservation.}
    \vspace{-1.5em}
    \label{tbl:re_id_and_user_study}
\end{table}

\subsection{Ablation Study}
\label{sec:ablation}
\vspace{-1mm}

We show several ablation studies to motivate the finetuning stage that injects the personalized prior and the choice of physical, pixel-aligned buffers to condition the model.
\vspace{-2mm}
\myparagraph{No Personalized Priors}
We first show how \model performs in the absence of personalized priors (\ie, trained on only the large dataset from Stage 1).
Figure~\ref{fig:reconstruction} shows that our model learns to use the physical buffers as conditions for pose, expression, and lighting, but it is incapable of preserving the person's identity during appearance editing.

\begin{figure}[ht]
\centering
\includegraphics[width=0.9\columnwidth]{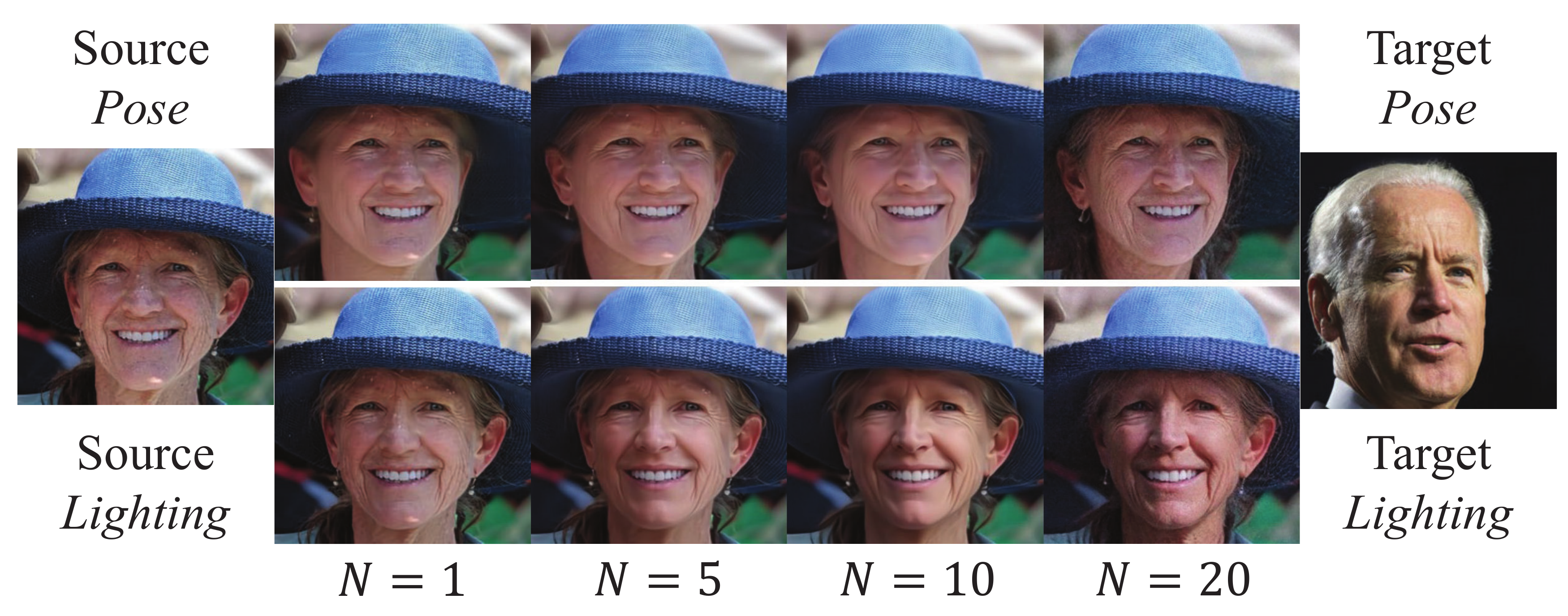}
\vspace{-.5em}
\caption{
\textbf{Quality \wrt number of Stage 2 images.}
\model achieves high-quality relighting and pose change with 20 images for Stage 2.
Using fewer may yield blurry results and make them hard to rig with new conditions.
}
\vspace{-1em}
\label{fig:num_images}
\end{figure}

\vspace{-2mm}
\myparagraph{Number of Images}
Here we explore how the number of images used in Stage 2 affects \model's ability of learning personalized priors.
We train three models of a non-celebrity with 1, 5, 10, and 20 images and test them on relighting, expression change, and pose change.
As Figure~\ref{fig:num_images} demonstrates, using just 1, 5, or 10 images yields worse results than using 20 images (unsurprisingly).
With more images, \model learns better-personalized priors that capture high-frequency face characteristics, such as the wrinkles in Figure~\ref{fig:num_images}.

\vspace{-2mm}

\begin{figure}[h]
\centering
\includegraphics[width=0.9\linewidth]{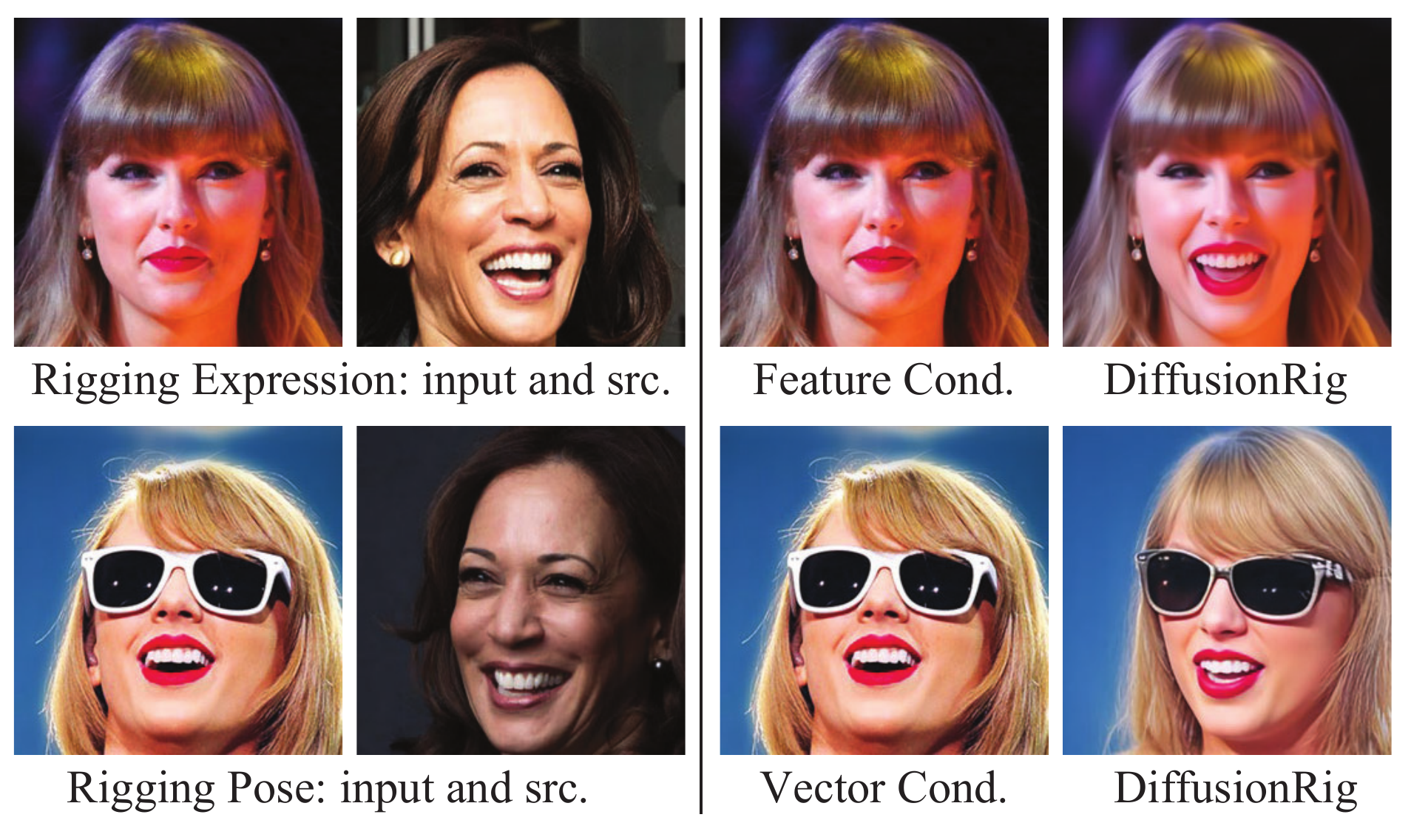}
\vspace{-3mm}
\caption{\textbf{Ablation on the form of conditions.}
Neither feature conditioning nor vector conditioning is able to rig the input image to follow the physical properties of the target image.}
\vspace{-5mm}
\label{fig:ablation}
\end{figure}

\myparagraph{Different Forms of Conditions}
\label{para:ab_cond}
There are alternative ways to condition the image synthesis.
We demonstrate that pixel-aligned physical buffers are the most effective form in accurately rigging the appearance.
We explore the following two conditioning alternatives.
\textbf{``Vector cond.''} is when we directly concatenate DECA parameters, a 236-dimensional vector, to the global latent code without using pixel-aligned buffers.
\textbf{``Feature cond.''} means that we concatenate the physical buffers to the input image and pass them into the encoder to compute a global latent code, which is then used as a non-spatial feature condition.
As shown in Figure~\ref{fig:ablation}, using pixel-aligned physical guidance is essential for accurate conditional image editing.
Both vector and feature conditioning suffer from the generated images not following the desired physical guidance. 

\vspace{-2mm}

\vspace{-1mm}
\section{Limitations \& Conclusion}
\vspace{-1mm}
Although \model achieves state-of-the-art facial appearance editing, it relies on a small portrait dataset to finetune, which limits its scalability for massive user adoption. 
Furthermore, when the edit involves dramatic head pose change, \model may not stay faithful to the original background, since head pose change sometimes reveals what used to be occluded, therefore requiring background inpainting---a topic beyond the scope of this paper.
Additionally, since \model relies on DECA to get physical buffers, it will also be affected by DECA's limited estimation capability: for instance, extreme expressions usually cannot be well predicted. and the estimated lighting is sometimes coupled with the skin tone.

In this paper, we have presented \model, a riggable diffusion model for identity-preserving, personalized editing of facial appearance.
We introduced a two-stage method to first learn generic face priors and later personalized priors.
Using both explicit conditioning via physical buffers and implicit conditioning via global latent code, we can drive and control our model's facial image synthesis.

\vspace{-2mm}
\myparagraph{Acknowledgment} We thank Marc Levoy for the valuable feedback and everyone whose photos appear in this paper for their permission.

{\small
\bibliographystyle{ieee_fullname}
\bibliography{egbib}

\begin{thebibliography}{10}\itemsep=-1pt

\bibitem{Bi_Lombardi_Saito_Simon_Wei_Mcphail_Ramamoorthi_Sheikh_Saragih}
Sai Bi, Stephen Lombardi, Shunsuke Saito, Tomas Simon, Shih-En Wei, Kevyn
  Mcphail, Ravi Ramamoorthi, Yaser Sheikh, and Jason Saragih.
\newblock Deep relightable appearance models for animatable faces.
\newblock 40(4):15.

\bibitem{blanz1999morphable}
Volker Blanz and Thomas Vetter.
\newblock A morphable model for the synthesis of 3d faces.
\newblock In {\em Proceedings of the 26th annual conference on Computer
  graphics and interactive techniques}, pages 187--194, 1999.

\bibitem{Chan_Lin_Chan_Nagano_Pan_DeMello_Gallo_Guibas_Tremblay_Khamis_etal_2022}
Eric~R. Chan, Connor~Z. Lin, Matthew~A. Chan, Koki Nagano, Boxiao Pan, Shalini
  De~Mello, Orazio Gallo, Leonidas Guibas, Jonathan Tremblay, Sameh Khamis,
  Tero Karras, and Gordon Wetzstein.
\newblock Efficient geometry-aware 3d generative adversarial networks.
\newblock (arXiv:2112.07945), Apr 2022.
\newblock arXiv:2112.07945 [cs].

\bibitem{chan2021pi}
Eric~R Chan, Marco Monteiro, Petr Kellnhofer, Jiajun Wu, and Gordon Wetzstein.
\newblock pi-gan: Periodic implicit generative adversarial networks for
  3d-aware image synthesis.
\newblock In {\em Proceedings of the IEEE/CVF conference on computer vision and
  pattern recognition}, pages 5799--5809, 2021.

\bibitem{choi2022perception}
Jooyoung Choi, Jungbeom Lee, Chaehun Shin, Sungwon Kim, Hyunwoo Kim, and
  Sungroh Yoon.
\newblock Perception prioritized training of diffusion models.
\newblock In {\em Proceedings of the IEEE/CVF Conference on Computer Vision and
  Pattern Recognition}, pages 11472--11481, 2022.

\bibitem{debevec2000acquiring}
Paul Debevec, Tim Hawkins, Chris Tchou, Haarm-Pieter Duiker, Westley Sarokin,
  and Mark Sagar.
\newblock Acquiring the reflectance field of a human face.
\newblock In {\em Proceedings of the 27th annual conference on Computer
  graphics and interactive techniques}, pages 145--156, 2000.

\bibitem{dhariwal2021diffusion}
Prafulla Dhariwal and Alexander Nichol.
\newblock Diffusion models beat gans on image synthesis.
\newblock {\em Advances in Neural Information Processing Systems},
  34:8780--8794, 2021.

\bibitem{Egger_Smith_Tewari_Wuhrer_Zollhoefer_Beeler_Bernard_Bolkart_Kortylewski_Romdhani_etal_2020}
Bernhard Egger, William A.~P. Smith, Ayush Tewari, Stefanie Wuhrer, Michael
  Zollhoefer, Thabo Beeler, Florian Bernard, Timo Bolkart, Adam Kortylewski,
  Sami Romdhani, Christian Theobalt, Volker Blanz, and Thomas Vetter.
\newblock 3d morphable face models -- past, present and future.
\newblock {\em arXiv:1909.01815 [cs]}, Apr 2020.
\newblock arXiv: 1909.01815.

\bibitem{feng2021learning}
Yao Feng, Haiwen Feng, Michael~J Black, and Timo Bolkart.
\newblock Learning an animatable detailed 3d face model from in-the-wild
  images.
\newblock {\em ACM Transactions on Graphics (ToG)}, 40(4):1--13, 2021.

\bibitem{ghosh2020gif}
Partha Ghosh, Pravir~Singh Gupta, Roy Uziel, Anurag Ranjan, Michael~J Black,
  and Timo Bolkart.
\newblock Gif: Generative interpretable faces.
\newblock In {\em 2020 International Conference on 3D Vision (3DV)}, pages
  868--878. IEEE, 2020.

\bibitem{goodfellow2020generative}
Ian Goodfellow, Jean Pouget-Abadie, Mehdi Mirza, Bing Xu, David Warde-Farley,
  Sherjil Ozair, Aaron Courville, and Yoshua Bengio.
\newblock Generative adversarial networks.
\newblock {\em Communications of the ACM}, 63(11):139--144, 2020.

\bibitem{grassal2022neural}
Philip-William Grassal, Malte Prinzler, Titus Leistner, Carsten Rother,
  Matthias Nie{\ss}ner, and Justus Thies.
\newblock Neural head avatars from monocular rgb videos.
\newblock In {\em Proceedings of the IEEE/CVF Conference on Computer Vision and
  Pattern Recognition}, pages 18653--18664, 2022.

\bibitem{Hao_Mallya_Belongie_Liu_2021}
Zekun Hao, Arun Mallya, Serge Belongie, and Ming-Yu Liu.
\newblock Gancraft: Unsupervised 3d neural rendering of minecraft worlds.
\newblock {\em arXiv:2104.07659 [cs]}, Apr 2021.
\newblock arXiv: 2104.07659.

\bibitem{he2016deep}
Kaiming He, Xiangyu Zhang, Shaoqing Ren, and Jian Sun.
\newblock Deep residual learning for image recognition.
\newblock In {\em Proceedings of the IEEE conference on computer vision and
  pattern recognition}, pages 770--778, 2016.

\bibitem{ho2020denoising}
Jonathan Ho, Ajay Jain, and Pieter Abbeel.
\newblock Denoising diffusion probabilistic models.
\newblock {\em Advances in Neural Information Processing Systems},
  33:6840--6851, 2020.

\bibitem{Ho_Jain_Abbeel_2020}
Jonathan Ho, Ajay Jain, and Pieter Abbeel.
\newblock Denoising diffusion probabilistic models.
\newblock In {\em Advances in Neural Information Processing Systems},
  volume~33, page 6840–6851. Curran Associates, Inc., 2020.

\bibitem{hong2021headnerf}
Yang Hong, Bo Peng, Haiyao Xiao, Ligang Liu, and Juyong Zhang.
\newblock Headnerf: A real-time nerf-based parametric head model.
\newblock In {\em {IEEE/CVF} Conference on Computer Vision and Pattern
  Recognition (CVPR)}, 2022.

\bibitem{karras2021alias}
Tero Karras, Miika Aittala, Samuli Laine, Erik H{\"a}rk{\"o}nen, Janne
  Hellsten, Jaakko Lehtinen, and Timo Aila.
\newblock Alias-free generative adversarial networks.
\newblock {\em Advances in Neural Information Processing Systems}, 34:852--863,
  2021.

\bibitem{karras2019style}
Tero Karras, Samuli Laine, and Timo Aila.
\newblock A style-based generator architecture for generative adversarial
  networks.
\newblock In {\em Proceedings of the IEEE/CVF conference on computer vision and
  pattern recognition}, pages 4401--4410, 2019.

\bibitem{Karras_Laine_Aila_2019}
Tero Karras, Samuli Laine, and Timo Aila.
\newblock A style-based generator architecture for generative adversarial
  networks.
\newblock (arXiv:1812.04948), Mar 2019.
\newblock arXiv:1812.04948 [cs, stat].

\bibitem{Karras_Laine_Aittala_Hellsten_Lehtinen_Aila_2020}
Tero Karras, Samuli Laine, Miika Aittala, Janne Hellsten, Jaakko Lehtinen, and
  Timo Aila.
\newblock Analyzing and improving the image quality of stylegan.
\newblock (arXiv:1912.04958), Mar 2020.
\newblock arXiv:1912.04958 [cs, eess, stat].

\bibitem{dlib09}
Davis~E. King.
\newblock Dlib-ml: A machine learning toolkit.
\newblock {\em Journal of Machine Learning Research}, 10:1755--1758, 2009.

\bibitem{kingma2014adam}
Diederik~P Kingma and Jimmy Ba.
\newblock Adam: A method for stochastic optimization.
\newblock {\em arXiv preprint arXiv:1412.6980}, 2014.

\bibitem{li2017learning}
Tianye Li, Timo Bolkart, Michael~J Black, Hao Li, and Javier Romero.
\newblock Learning a model of facial shape and expression from 4d scans.
\newblock {\em ACM Trans. Graph.}, 36(6):194--1, 2017.

\bibitem{li2020blind}
Xiaoming Li, Chaofeng Chen, Shangchen Zhou, Xianhui Lin, Wangmeng Zuo, and Lei
  Zhang.
\newblock Blind face restoration via deep multi-scale component dictionaries.
\newblock In {\em European Conference on Computer Vision}, pages 399--415.
  Springer, 2020.

\bibitem{liu20223d}
Yuchen Liu, Zhixin Shu, Yijun Li, Zhe Lin, Richard Zhang, and SY Kung.
\newblock 3d-fm gan: Towards 3d-controllable face manipulation.
\newblock In {\em European Conference on Computer Vision}, pages 107--125.
  Springer, 2022.

\bibitem{liu2015faceattributes}
Ziwei Liu, Ping Luo, Xiaogang Wang, and Xiaoou Tang.
\newblock Deep learning face attributes in the wild.
\newblock In {\em Proceedings of International Conference on Computer Vision
  (ICCV)}, December 2015.

\bibitem{Lombardi_Saragih_Simon_Sheikh_2018}
Stephen Lombardi, Jason Saragih, Tomas Simon, and Yaser Sheikh.
\newblock Deep appearance models for face rendering.
\newblock {\em ACM Transactions on Graphics}, 37(4):1–13, Jul 2018.
\newblock arXiv: 1808.00362.

\bibitem{meka2019deep}
Abhimitra Meka, Christian Haene, Rohit Pandey, Michael Zollh{\"o}fer, Sean
  Fanello, Graham Fyffe, Adarsh Kowdle, Xueming Yu, Jay Busch, Jason
  Dourgarian, et~al.
\newblock Deep reflectance fields: high-quality facial reflectance field
  inference from color gradient illumination.
\newblock {\em ACM Transactions on Graphics (TOG)}, 38(4):1--12, 2019.

\bibitem{nestmeyer2020learning}
Thomas Nestmeyer, Jean-Fran{\c{c}}ois Lalonde, Iain Matthews, and Andreas
  Lehrmann.
\newblock Learning physics-guided face relighting under directional light.
\newblock In {\em Proceedings of the IEEE/CVF Conference on Computer Vision and
  Pattern Recognition}, pages 5124--5133, 2020.

\bibitem{nichol2021improved}
Alexander~Quinn Nichol and Prafulla Dhariwal.
\newblock Improved denoising diffusion probabilistic models.
\newblock In {\em International Conference on Machine Learning}, pages
  8162--8171. PMLR, 2021.

\bibitem{nitzan2022mystyle}
Yotam Nitzan, Kfir Aberman, Qiurui He, Orly Liba, Michal Yarom, Yossi
  Gandelsman, Inbar Mosseri, Yael Pritch, and Daniel Cohen-Or.
\newblock Mystyle: A personalized generative prior.
\newblock {\em arXiv preprint arXiv:2203.17272}, 2022.

\bibitem{preechakul2022diffusion}
Konpat Preechakul, Nattanat Chatthee, Suttisak Wizadwongsa, and Supasorn
  Suwajanakorn.
\newblock Diffusion autoencoders: Toward a meaningful and decodable
  representation.
\newblock In {\em Proceedings of the IEEE/CVF Conference on Computer Vision and
  Pattern Recognition}, pages 10619--10629, 2022.

\bibitem{Preechakul_Chatthee_Wizadwongsa_Suwajanakorn_2022}
Konpat Preechakul, Nattanat Chatthee, Suttisak Wizadwongsa, and Supasorn
  Suwajanakorn.
\newblock Diffusion autoencoders: Toward a meaningful and decodable
  representation.
\newblock {\em arXiv:2111.15640 [cs]}, Mar 2022.
\newblock arXiv: 2111.15640.

\bibitem{R_Tewari_Dib_Weyrich_Bickel_Seidel_Pfister_Matusik_Chevallier_Elgharib_etal_2021}
Mallikarjun~B. R, Ayush Tewari, Abdallah Dib, Tim Weyrich, Bernd Bickel,
  Hans-Peter Seidel, Hanspeter Pfister, Wojciech Matusik, Louis Chevallier,
  Mohamed Elgharib, and Christian Theobalt.
\newblock Photoapp: Photorealistic appearance editing of head portraits.
\newblock {\em arXiv:2103.07658 [cs]}, Mar 2021.
\newblock arXiv: 2103.07658.

\bibitem{remelli2022drivable}
Edoardo Remelli, Timur Bagautdinov, Shunsuke Saito, Chenglei Wu, Tomas Simon,
  Shih-En Wei, Kaiwen Guo, Zhe Cao, Fabian Prada, Jason Saragih, et~al.
\newblock Drivable volumetric avatars using texel-aligned features.
\newblock In {\em ACM SIGGRAPH 2022 Conference Proceedings}, pages 1--9, 2022.

\bibitem{saharia2022image}
Chitwan Saharia, Jonathan Ho, William Chan, Tim Salimans, David~J Fleet, and
  Mohammad Norouzi.
\newblock Image super-resolution via iterative refinement.
\newblock {\em IEEE Transactions on Pattern Analysis and Machine Intelligence},
  2022.

\bibitem{Sanyal_Bolkart_Feng_Black_2019}
Soubhik Sanyal, Timo Bolkart, Haiwen Feng, and Michael~J. Black.
\newblock Learning to regress 3d face shape and expression from an image
  without 3d supervision.
\newblock (arXiv:1905.06817), May 2019.
\newblock arXiv:1905.06817 [cs].

\bibitem{Sengupta_Kanazawa_Castillo_Jacobs_2018}
Soumyadip Sengupta, Angjoo Kanazawa, Carlos~D. Castillo, and David Jacobs.
\newblock Sfsnet: Learning shape, reflectance and illuminance of faces in the
  wild.
\newblock {\em arXiv:1712.01261 [cs]}, Apr 2018.
\newblock arXiv: 1712.01261.

\bibitem{Shysheya_Zakharov_Aliev_Bashirov_Burkov_Iskakov_Ivakhnenko_Malkov_Pasechnik_Ulyanov_etal}
Aliaksandra Shysheya, Egor Zakharov, Kara-Ali Aliev, Renat Bashirov, Egor
  Burkov, Karim Iskakov, Aleksei Ivakhnenko, Yury Malkov, Igor Pasechnik,
  Dmitry Ulyanov, Alexander Vakhitov, and Victor Lempitsky.
\newblock Textured neural avatars.
\newblock page~11.

\bibitem{song2020denoising}
Jiaming Song, Chenlin Meng, and Stefano Ermon.
\newblock Denoising diffusion implicit models.
\newblock {\em arXiv preprint arXiv:2010.02502}, 2020.

\bibitem{sun2019single}
Tiancheng Sun, Jonathan~T Barron, Yun-Ta Tsai, Zexiang Xu, Xueming Yu, Graham
  Fyffe, Christoph Rhemann, Jay Busch, Paul~E Debevec, and Ravi Ramamoorthi.
\newblock Single image portrait relighting.
\newblock {\em ACM Trans. Graph.}, 38(4):79--1, 2019.

\bibitem{Tan_Fanello_Meka_Orts-Escolano_Tang_Pandey_Taylor_Tan_Zhang_2022}
Feitong Tan, Sean Fanello, Abhimitra Meka, Sergio Orts-Escolano, Danhang Tang,
  Rohit Pandey, Jonathan Taylor, Ping Tan, and Yinda Zhang.
\newblock Volux-gan: A generative model for 3d face synthesis with hdri
  relighting.
\newblock (arXiv:2201.04873), Jan 2022.
\newblock number: arXiv:2201.04873 arXiv:2201.04873 [cs].

\bibitem{tewari2020pie}
Ayush Tewari, Mohamed Elgharib, Florian Bernard, Hans-Peter Seidel, Patrick
  P{\'e}rez, Michael Zollh{\"o}fer, and Christian Theobalt.
\newblock Pie: Portrait image embedding for semantic control.
\newblock {\em ACM Transactions on Graphics (TOG)}, 39(6):1--14, 2020.

\bibitem{tewari2020stylerig}
Ayush Tewari, Mohamed Elgharib, Gaurav Bharaj, Florian Bernard, Hans-Peter
  Seidel, Patrick P{\'e}rez, Michael Zollhofer, and Christian Theobalt.
\newblock Stylerig: Rigging stylegan for 3d control over portrait images.
\newblock In {\em Proceedings of the IEEE/CVF Conference on Computer Vision and
  Pattern Recognition}, pages 6142--6151, 2020.

\bibitem{tewari2022disentangled3d}
Ayush Tewari, Xingang Pan, Ohad Fried, Maneesh Agrawala, Christian Theobalt,
  et~al.
\newblock Disentangled3d: Learning a 3d generative model with disentangled
  geometry and appearance from monocular images.
\newblock In {\em Proceedings of the IEEE/CVF Conference on Computer Vision and
  Pattern Recognition}, pages 1516--1525, 2022.

\bibitem{wang2020multiple}
Kaili Wang, Jose Oramas, and Tinne Tuytelaars.
\newblock Multiple exemplars-based hallucination for face super-resolution and
  editing.
\newblock In {\em Proceedings of the Asian Conference on Computer Vision},
  2020.

\bibitem{Wang_Chen_Yu_Ma_Li_Liu_2022}
Lizhen Wang, Zhiyuan Chen, Tao Yu, Chenguang Ma, Liang Li, and Yebin Liu.
\newblock Faceverse: a fine-grained and detail-controllable 3d face morphable
  model from a hybrid dataset.
\newblock (arXiv:2203.14057), May 2022.
\newblock arXiv:2203.14057 [cs].

\bibitem{wang2022restoreformer}
Zhouxia Wang, Jiawei Zhang, Runjian Chen, Wenping Wang, and Ping Luo.
\newblock Restoreformer: High-quality blind face restoration from undegraded
  key-value pairs.
\newblock In {\em Proceedings of the IEEE/CVF Conference on Computer Vision and
  Pattern Recognition}, pages 17512--17521, 2022.

\bibitem{Wuu_Zheng_Ardisson_Bali_Belko_Brockmeyer_Evans_Godisart_Ha_Hypes_etal_2022}
Cheng-hsin Wuu, Ningyuan Zheng, Scott Ardisson, Rohan Bali, Danielle Belko,
  Eric Brockmeyer, Lucas Evans, Timothy Godisart, Hyowon Ha, Alexander Hypes,
  Taylor Koska, Steven Krenn, Stephen Lombardi, Xiaomin Luo, Kevyn McPhail,
  Laura Millerschoen, Michal Perdoch, Mark Pitts, Alexander Richard, Jason
  Saragih, Junko Saragih, Takaaki Shiratori, Tomas Simon, Matt Stewart, Autumn
  Trimble, Xinshuo Weng, David Whitewolf, Chenglei Wu, Shoou-I. Yu, and Yaser
  Sheikh.
\newblock Multiface: A dataset for neural face rendering.
\newblock (arXiv:2207.11243), Jul 2022.
\newblock arXiv:2207.11243 [cs].

\bibitem{yu2018super}
Xin Yu, Basura Fernando, Richard Hartley, and Fatih Porikli.
\newblock Super-resolving very low-resolution face images with supplementary
  attributes.
\newblock In {\em Proceedings of the IEEE conference on computer vision and
  pattern recognition}, pages 908--917, 2018.

\bibitem{Zehni_Ghosh_Sridhar_Raman_2021}
Mona Zehni, Shaona Ghosh, Krishna Sridhar, and Sethu Raman.
\newblock Joint learning of portrait intrinsic decomposition and relighting.
\newblock {\em arXiv:2106.15305 [cs]}, Jun 2021.
\newblock arXiv: 2106.15305.

\bibitem{zhang2020portrait}
Xuaner Zhang, Jonathan~T Barron, Yun-Ta Tsai, Rohit Pandey, Xiuming Zhang, Ren
  Ng, and David~E Jacobs.
\newblock Portrait shadow manipulation.
\newblock {\em ACM Transactions on Graphics (TOG)}, 39(4):78--1, 2020.

\bibitem{zhao2022generative}
Xiaoming Zhao, Fangchang Ma, David G{\"u}era, Zhile Ren, Alexander~G Schwing,
  and Alex Colburn.
\newblock Generative multiplane images: Making a 2d gan 3d-aware.
\newblock In {\em European Conference on Computer Vision}, pages 18--35.
  Springer, 2022.

\end{thebibliography}
}

\clearpage

\pagenumbering{arabic}
\renewcommand*{\thepage}{A\arabic{page}}

\setcounter{figure}{0}
\renewcommand\thefigure{S\arabic{figure}}   

\setcounter{table}{0}
\renewcommand\thetable{S\arabic{table}}   

\newcommand{\appendixhead}
{\centering{\Large \bf Supplementary Material}
\vspace{10mm}}

\twocolumn[\appendixhead]

\appendix

\section{More Results on Personalized Editing}

\begin{figure}[h!]
    \centering
    \setlength{\tabcolsep}{2pt}
    \def\imW{0.32\linewidth}
    \begin{tabular}{ccc}
        \includegraphics[width=\imW]{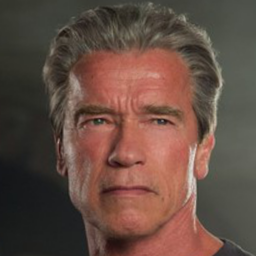} &
        \includegraphics[width=\imW]{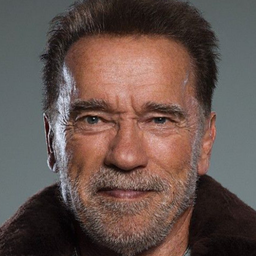} & 
        \includegraphics[width=\imW]{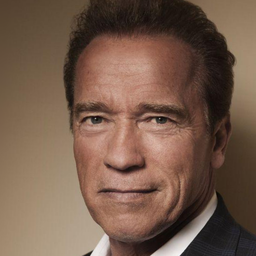} \\
        \hline\vspace{-8pt}\\
        \includegraphics[width=\imW]{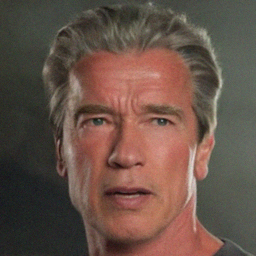} & 
        \includegraphics[width=\imW]{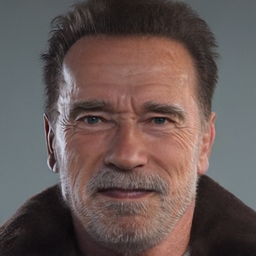} & 
        \includegraphics[width=\imW]{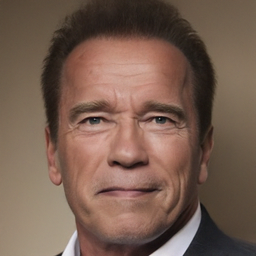} \\
        \includegraphics[width=\imW]{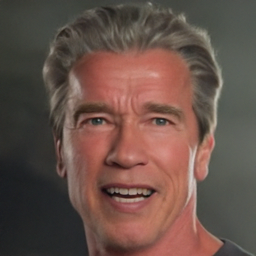} &
        \includegraphics[width=\imW]{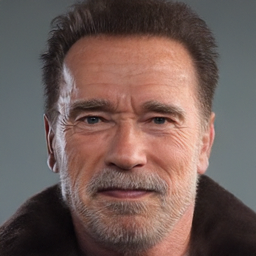} &
        \includegraphics[width=\imW]{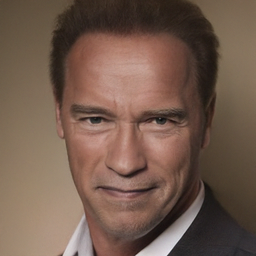} \\
        \includegraphics[width=\imW]{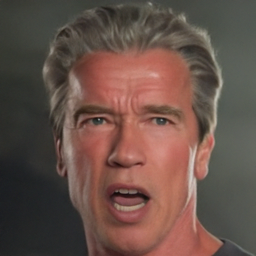} &
        \includegraphics[width=\imW]{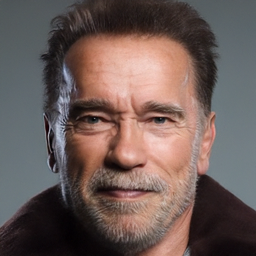} &
        \includegraphics[width=\imW]{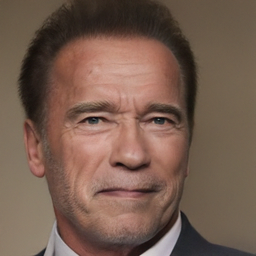} \\
        \includegraphics[width=\imW]{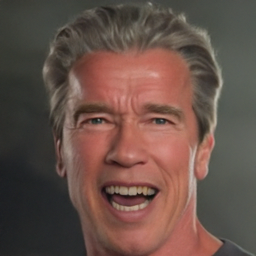} &
        \includegraphics[width=\imW]{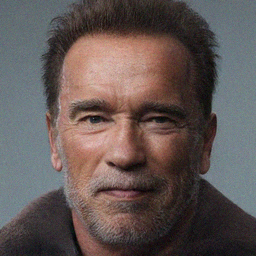} &
        \includegraphics[width=\imW]{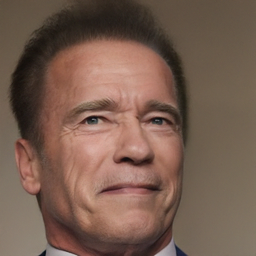} \\
        {\footnotesize Expression} & {\footnotesize Lighting} & {\footnotesize Pose}
    \end{tabular}
    \caption{\textbf{More results on using physical buffers to rig the facial appearance.}
    The physical buffers (not shown) are used to edit the input images (top row) in terms of facial expression (left column), lighting (middle column), and head pose (right column).
    }
    \label{fig:appendA}
\end{figure}

We provide more results on using physical buffers to rig/drive facial appearance generation in Figure \ref{fig:appendA}.

\section{Extreme Lighting Editing}

\begin{figure}[ht]
\centering
\setlength{\tabcolsep}{1.0pt}
\def\imW{0.33\linewidth}
\begin{tabular}{ccc}
\includegraphics[width=\imW]{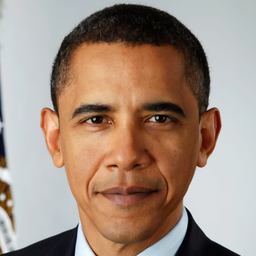}
& 
\includegraphics[width=\imW]{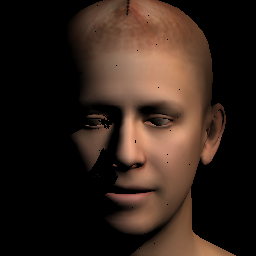}
&
\includegraphics[width=\imW]{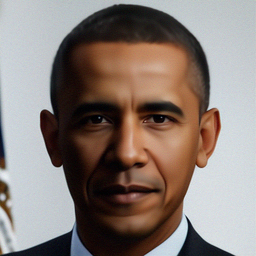}
\\
\includegraphics[width=\imW]{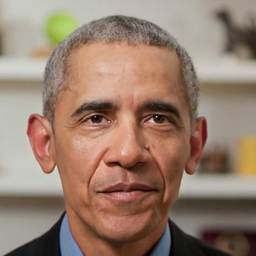}
&
\includegraphics[width=\imW]{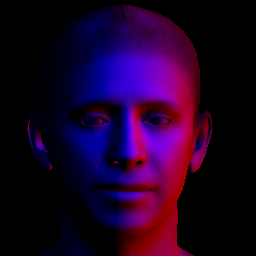}
& 
\includegraphics[width=\imW]{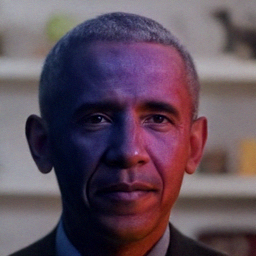}
\\
{\footnotesize Input} & {\footnotesize Target Lighting} & {\footnotesize Output} 
\end{tabular}
\caption{Stress test with difficult directional and RGB lighting.
}
\label{fig:dir_light}

\end{figure}

During the training of \model, we rely on the SH-based lighting model from DECA, which is limited in modeling high-frequency lighting. 
At inference time, we can use a different lighting representation that can model directional lighting with cast shadow (through ray casting).
We show one such extreme lighting example and another RGB lighting example in Figure~\ref{fig:dir_light}, for which our model 
regresses slightly towards less extreme lighting but still produces reasonable results.

\section{Personal Photo Collections}

\begin{figure*}[ht]
  \centering
  \includegraphics[width=0.95\textwidth]{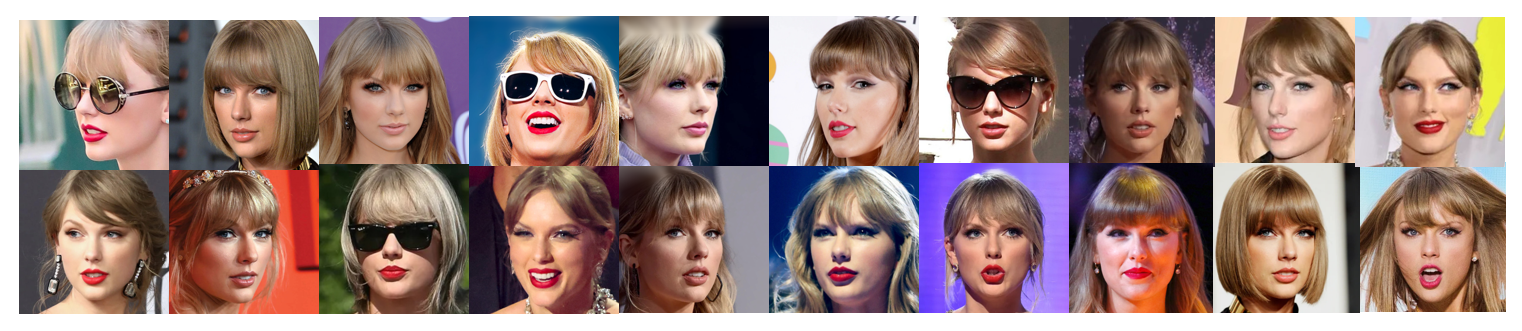}
  \\ 
  \includegraphics[width=0.945\textwidth]{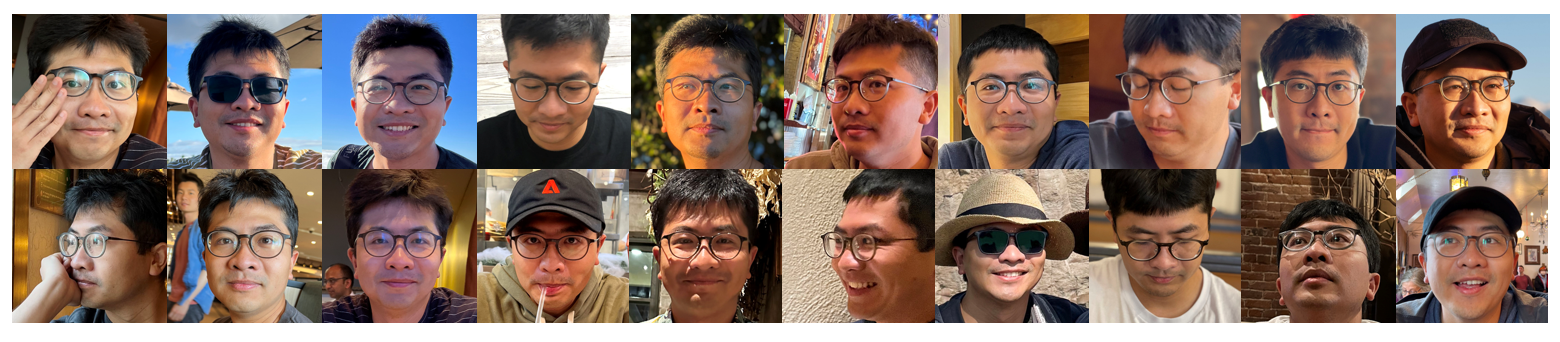}
   \caption{Personal photo collections used for training Stage 2: Taylor Swift and a non-celebrity.}
   \label{fig:celeb}
   \vspace{10mm}
\end{figure*}

In Figure~\ref{fig:celeb}, we show two sets of images we used to train Stage 2. 
For celebrities, we crawl the photos from the internet; for non-celebrities, we use everyday photos.
In comparison, MyStyle requires 92--279 images for finetuning.
When using only 20 images as we do, MyStyle cannot learn personalized priors well as shown in the main paper.

\begin{figure*}[h!]
\centering
\setlength{\tabcolsep}{0.5pt}
\def\imW{0.13\linewidth}
\begin{tabular}{ccccccc}
\includegraphics[width=\imW]{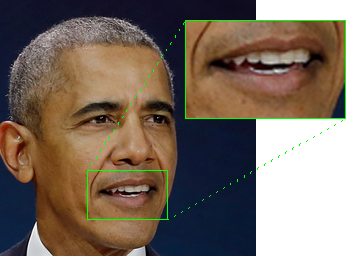}
& 
\includegraphics[width=\imW]{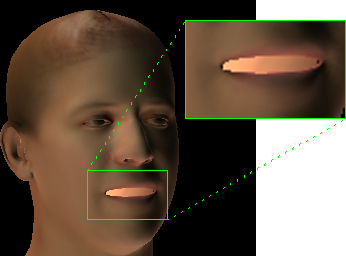}
&
\includegraphics[width=\imW]{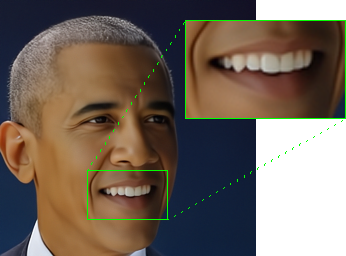}
&
\includegraphics[width=\imW]{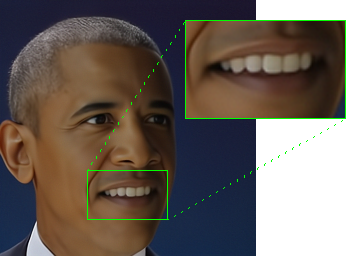}
&
\includegraphics[width=\imW]{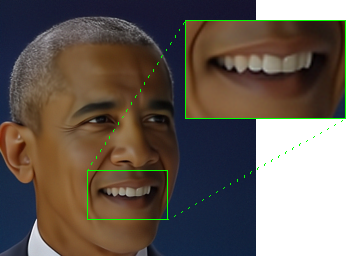}
&
\includegraphics[width=\imW]{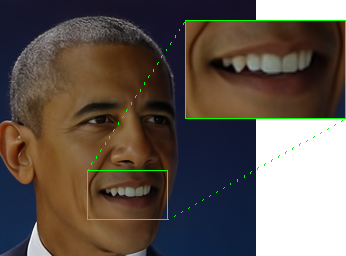}
& 
\includegraphics[width=\imW]{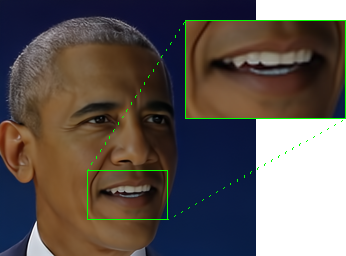}
\\
{\footnotesize Input} & {\footnotesize Target} & {\footnotesize N+S} & {\footnotesize L} & {\footnotesize N+L} & {\footnotesize A+L} & {\footnotesize A+N+L} \vspace{-1.5mm}\\
& {\footnotesize Lighting} & & & & & {\footnotesize (Ours)}
\end{tabular}
\caption{\textbf{Ablation on different input conditions.}
N: Normals, S: SH layers, L: Lambertian rendering, A: Albedo.
}
\label{fig:ab_diff_cond}

\end{figure*}

\section{Neural Network Architecture Details}

For our global encoder, we modify the ResNet-18 model by replacing the final classification layer with a feature extraction layer.
More specifically, we change the last layer into a linear layer that outputs our latent code.

Our diffusion model is based on the architecture presented in Guided-Diffusion \cite{dhariwal2021diffusion}.
We modify the architecture so that the model can take the global latent code as another condition (in addition to the concatenation of physical buffers and the noise image).
This global latent code is used for scaling and shifting the features.
Our model architecture details can be found in Table \ref{tab:arch}, where we provide hyperparameters for both our $256\times 256$ and $512\times 512$ models.

\begin{table}[ht]
    \vspace{2mm}
    \centering
    \scalebox{0.9}{
    \begin{tabular}{l|c|c}
    \toprule
        & $256\times 256$ & $512\times 512$ \\
        \hline 
    Diffusion Steps & $1000$ & $1000$ \\
    Channels & $128$ & $128$ \\
    Channels Multiple & $1,1,2,2,4,4$ & $0.5,1,1,2,2,4,4$ \\
    Heads Channels & $128$ & $128$ \\
    Attention Resolution & $16$ & $16$ \\
    Dropout & $0.1$ & $0.1$ \\
    P2\_gamma\textdagger & $1.0$ & $1.0$ \\
    P2\_k\textdagger & $1.0$ & $1.0$ \\
    Optimizer & Adam & Adam \\
    Weight Decay & $0.0$ & $0.0$ \\
    Batch Size (S1) & $256$ & $64$ \\
    Batch Size (S2) & $4$ & $2$ \\
    Iterations (S1) & $50$k & $200$k \\
    Iterations (S1) & $5$k & $20$k \\
    Learning Rate (S1) & $10^{-4}$ & $10^{-4}$ \\
    Learning Rate (S2) & $10^{-5}$ & $10^{-5}$\\
    \bottomrule
    \end{tabular}}
    \vspace{-2mm}
    \caption{\textbf{\model architecture details}.
    S1 and S2 denote Stages 1 and 2, respectively.
    Refer to Guided-Diffusion \cite{dhariwal2021diffusion} for more details.
    \textdagger\ are two hyperparamters defined in prior work \cite{choi2022perception}.}
    \label{tab:arch}
    \vspace{-2mm}
\end{table}

\section{Different Types of Pixel-Aligned Buffers}
We ablate different pixel-aligned buffers in Figure \ref{fig:ab_diff_cond}.  In our method, we use three kinds of physical buffers from DECA which are Normals (N), Albedo (A) and Lambertian rendering (L). With Lambertian rendering being the only physical buffer that contains lighting information, we include it in all our ablation studies except for the ``N+S'' where we use Normals and Spherical Harmonics with SH rendered on all-white albedo (\ie, shading), so it doesn't contain albedo information.
We can see with Normals, Albedo, and Lambertian rendering, the results preserve details (\eg, mouth) better, while N+S cannot render accurate lighting due to the missing albedo.

\section{Higher-Resolution Results ($512\times 512$)}

\begin{figure*}[h!]
    \centering
    \setlength{\tabcolsep}{2pt}
    \def\imW{0.24\linewidth}
    \begin{tabular}{cccc}
        &
        \includegraphics[width=\imW]{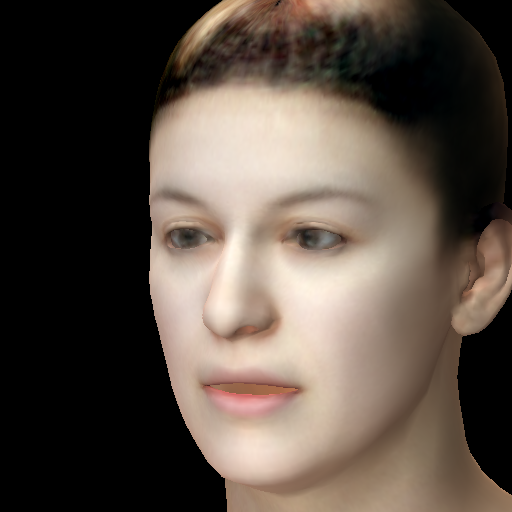} &
        \includegraphics[width=\imW]{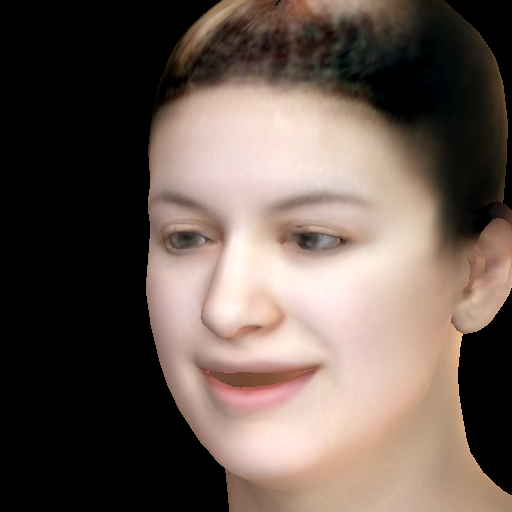} & 
        \includegraphics[width=\imW]{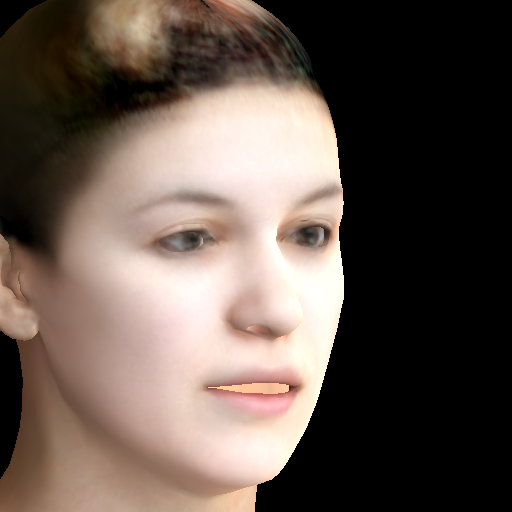} 
        \\
        \includegraphics[width=\imW]{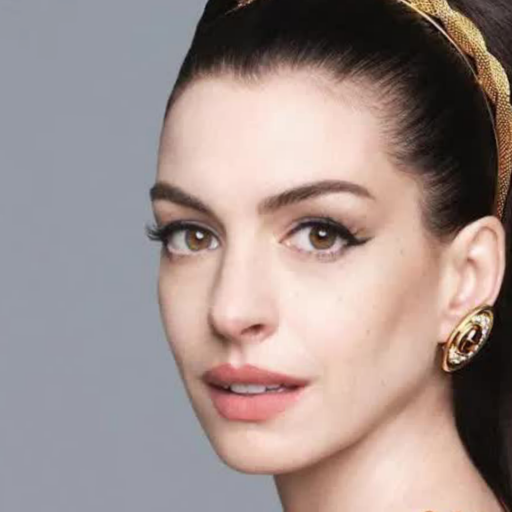} &
        \includegraphics[width=\imW]{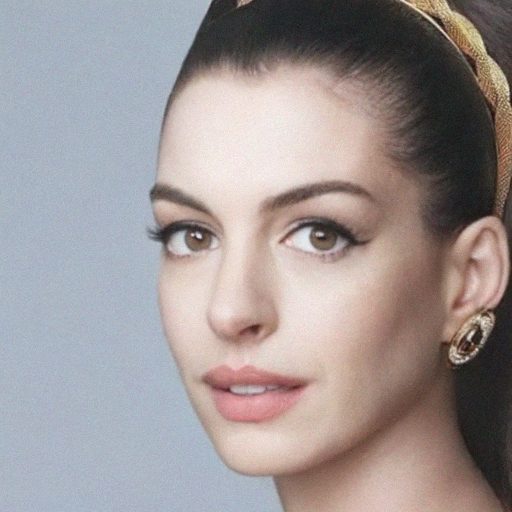} &
        \includegraphics[width=\imW]{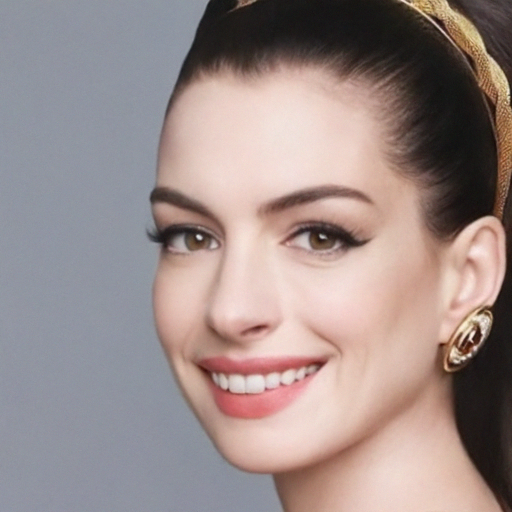} & 
        \includegraphics[width=\imW]{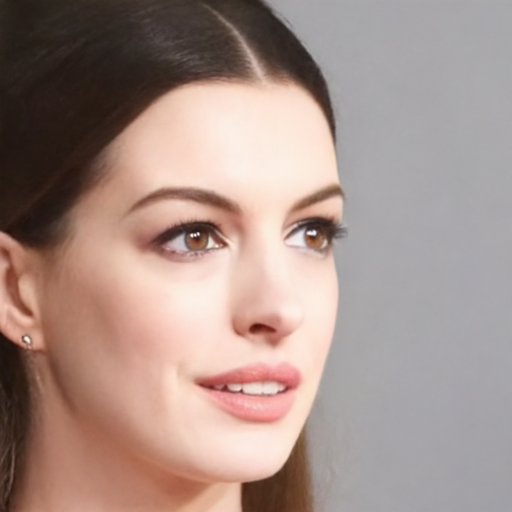} \\
        \midrule\vspace{-8pt}\\
        &
        \includegraphics[width=\imW]{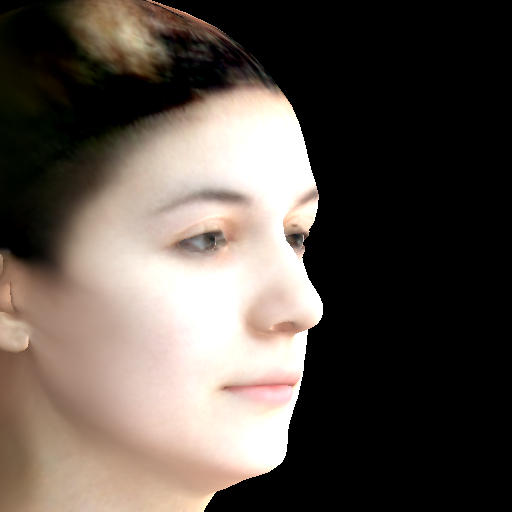} &
        \includegraphics[width=\imW]{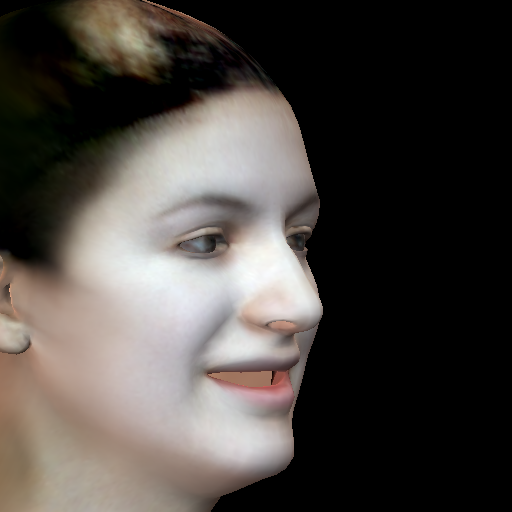} & 
        \includegraphics[width=\imW]{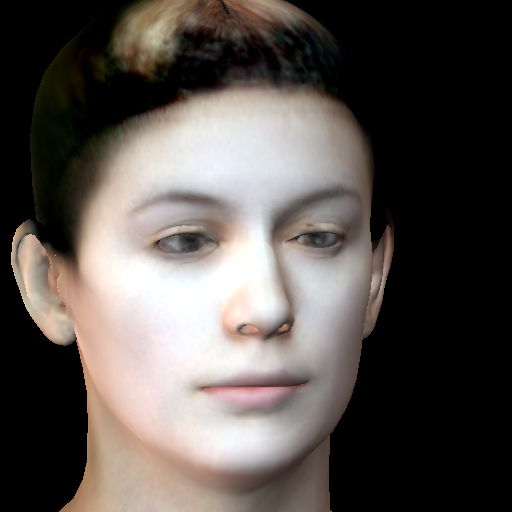} 
        \\
        \includegraphics[width=\imW]{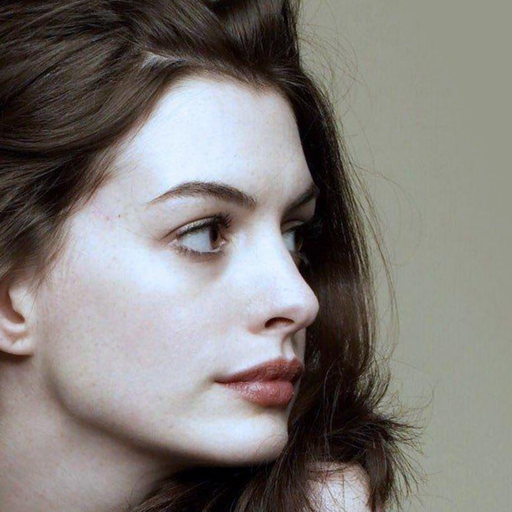} &
        \includegraphics[width=\imW]{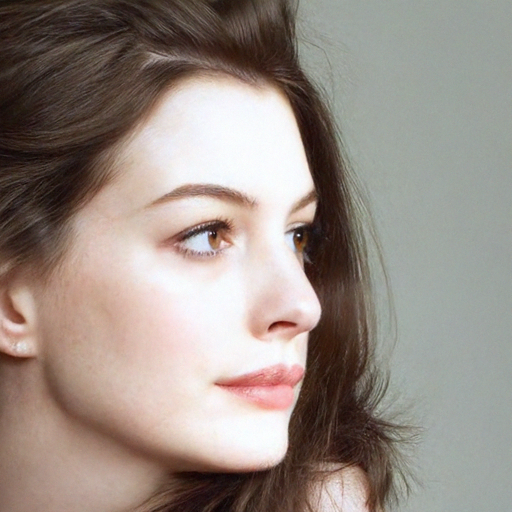} &
        \includegraphics[width=\imW]{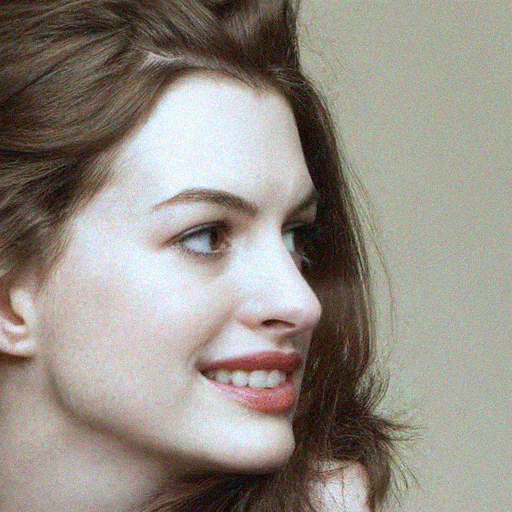} & 
        \includegraphics[width=\imW]{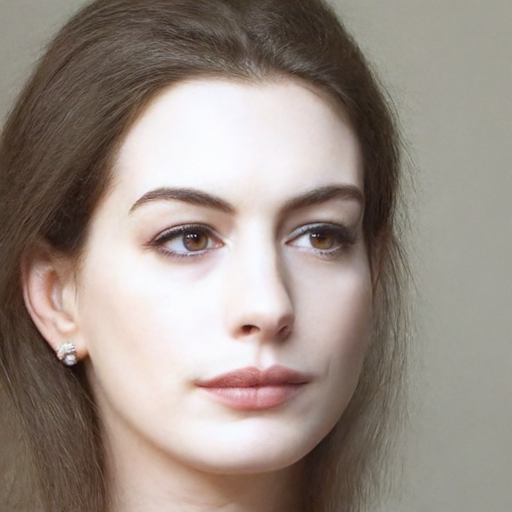} \\
        Input & Light & Expression & Pose
    \end{tabular}
    \caption{\textbf{512$\times$512 Facial Appearance Editing Results.} Two groups of results are presented here with the first row of each group being the physical buffers that drive the editing.}
    \vspace{25mm}
    \label{fig:512_1}
\end{figure*}

\begin{figure*}[h!]
    \centering
    \setlength{\tabcolsep}{2pt}
    \def\imW{0.24\linewidth}
    \begin{tabular}{cccc}
        &
        \includegraphics[width=\imW]{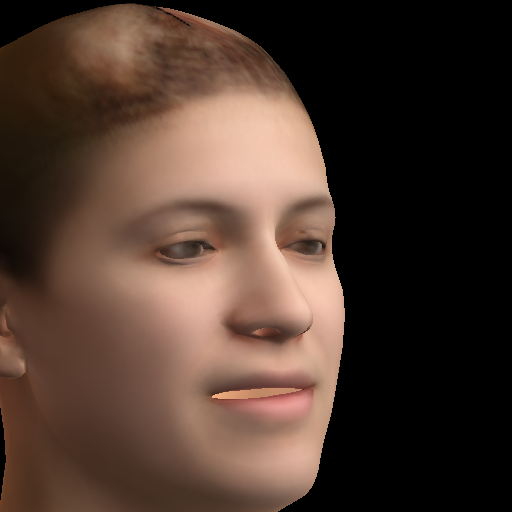} &
        \includegraphics[width=\imW]{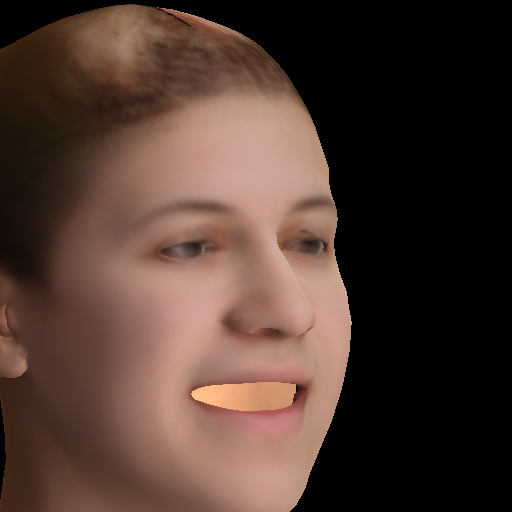} & 
        \includegraphics[width=\imW]{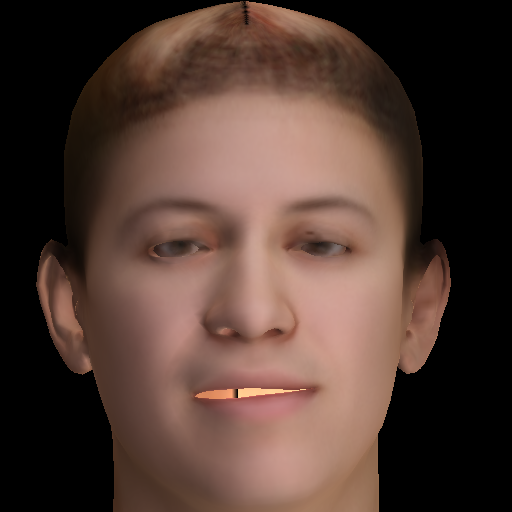} 
        \\
        \includegraphics[width=\imW]{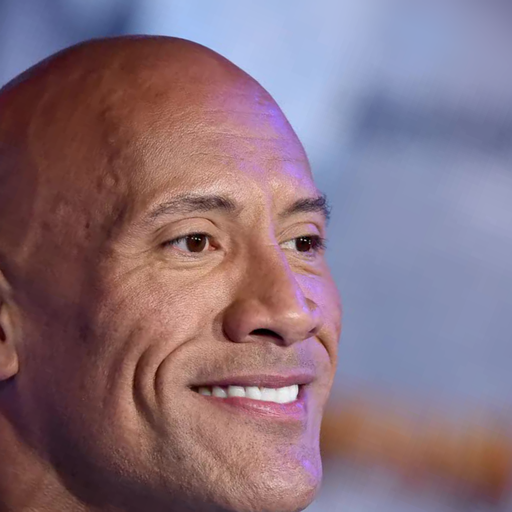} &
        \includegraphics[width=\imW]{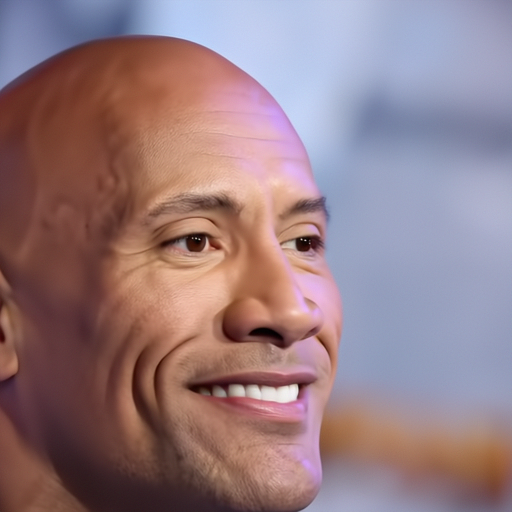} &
        \includegraphics[width=\imW]{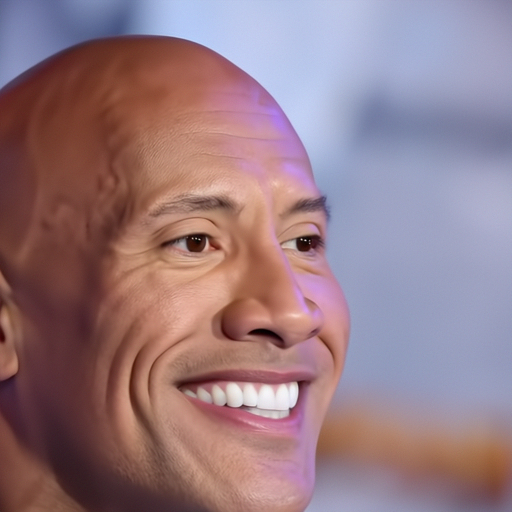} & 
        \includegraphics[width=\imW]{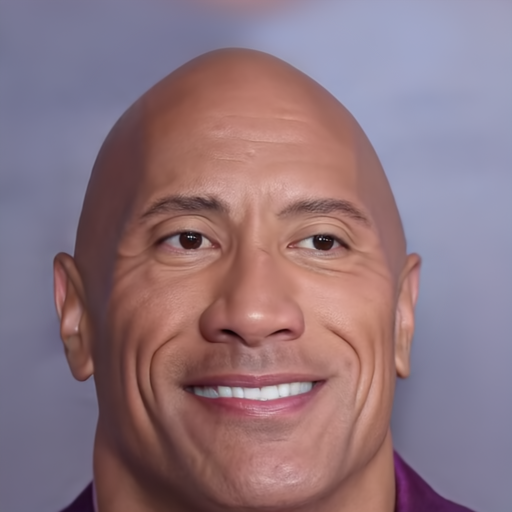} \\
        \midrule\vspace{-8pt}\\
        &
        \includegraphics[width=\imW]{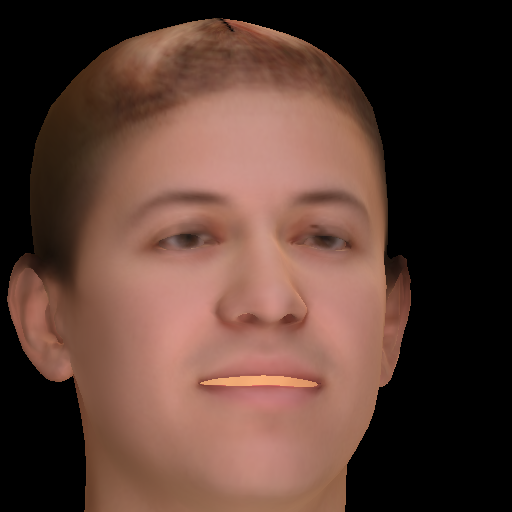} & 
        \includegraphics[width=\imW]{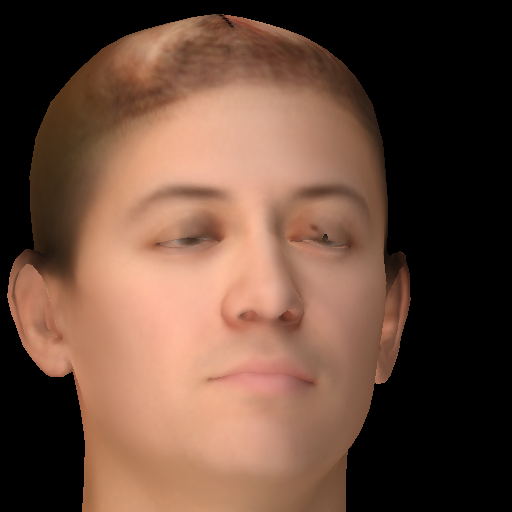} &
        \includegraphics[width=\imW]{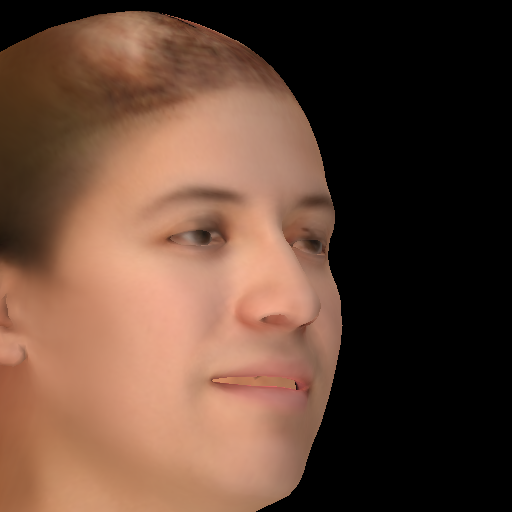} 
        \\
        \includegraphics[width=\imW]{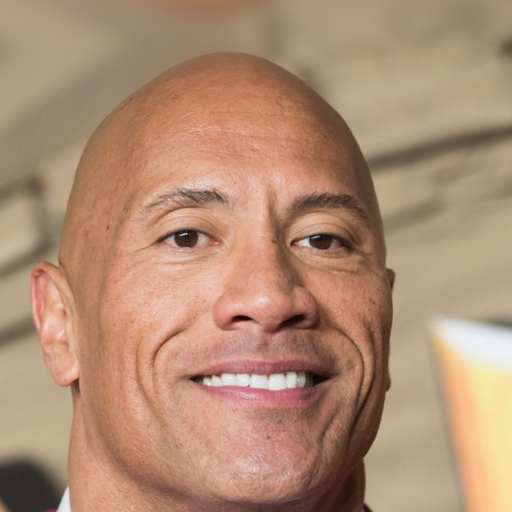} &
        \includegraphics[width=\imW]{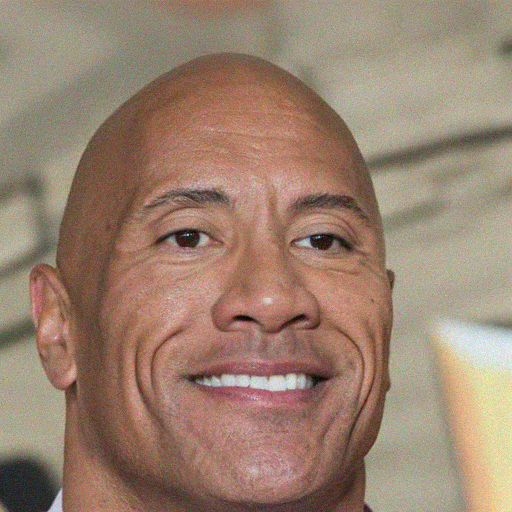} &
        \includegraphics[width=\imW]{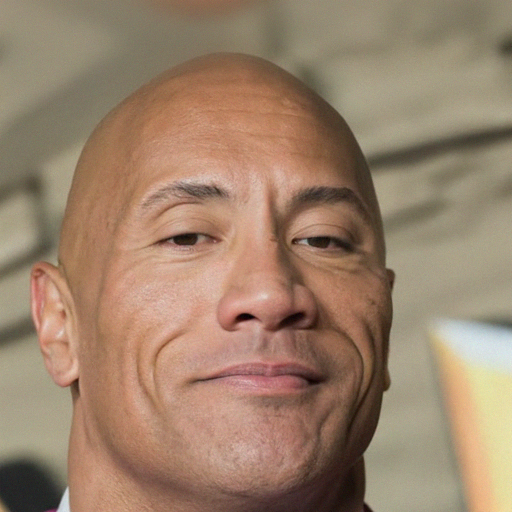} & 
        \includegraphics[width=\imW]{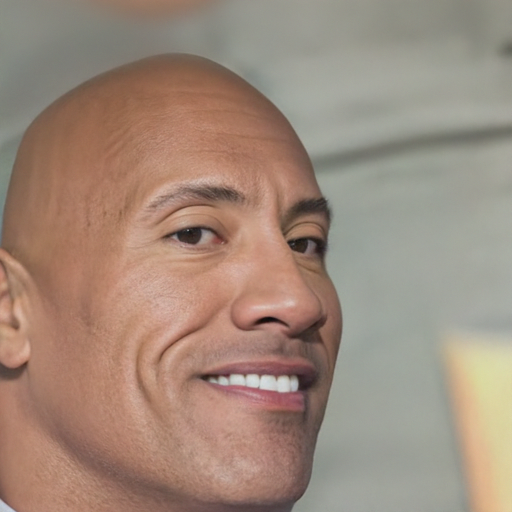} \\
        Input & Light & Expression & Pose
    \end{tabular}
    \caption{\textbf{512$\times$512 Facial Appearance Editing Results.} Two groups of results are presented here with the first row of each group being the physical buffers that drive the editing.}
    \label{fig:512_2}
    \vspace{30mm}
\end{figure*}

\model can be trained at $512\times 512$ resolution.
We show these higher-resolution results in Figures \ref{fig:512_1} and \ref{fig:512_2} on two new celebrities.

\end{document}